\documentclass[11pt,a4paper]{article}

\usepackage{acl}

\usepackage{times}
\usepackage{latexsym}
\usepackage{url}
\usepackage{amsmath}
\usepackage{graphicx}
\usepackage{subcaption}
\usepackage{xcolor}
\usepackage{booktabs}
\usepackage{colortbl}
\usepackage{makecell}
\usepackage{enumitem}
\usepackage{placeins}
\usepackage{hyperref}
\usepackage{fontawesome5}
\usepackage{simpleicons}
\usepackage{amssymb}
\usepackage[table]{xcolor}
\definecolor{PosGreen}{RGB}{0,140,80}
\definecolor{NegRed}{RGB}{200,40,40}

\title{PolyFact: Comparing Consistency-Driven Post-Training Methods for Cross-Lingual Factual Recall}

\author{
\begin{tabular}{c}
\textbf{Jonathan von Rad\thanks{Corresponding author.}  \quad Louis Arts \quad George Burgess \quad Eleftheria Kolokytha} \\
\textbf{Harry O'Donnell \quad Ektor Oikonomidis Doumpas \quad Eduardo Sánchez} \\
\textbf{Yao Lu \quad Pontus Stenetorp} \\[3pt]
\textnormal{University College London, Centre for Artificial Intelligence} \\[2pt]
{\normalsize\footnotesize\textnormal{\texttt{\{jonathan.rad.25,eduardo.sanchez,yao.lu\}@ucl.ac.uk}}}
\end{tabular}
}

\date{}

\begin{document}

\maketitle

\begin{abstract}
Large language models (LLMs) trained predominantly on English data encode substantial world knowledge, yet often fail to express it reliably in other languages, known as \emph{cross-lingual factual inconsistency}. To study, we introduce \textsc{PolyFact}, a fully parallel multilingual factual QA dataset of 60K Wikidata-grounded facts across 12 typologically diverse languages, and propose consistency-driven GRPO with \emph{cross-lingual reward pooling}. We compare our method against supervised fine-tuning (SFT) and the consistency-enhancement baselines DCO and CM-Align on OLMo-2-1124-7B and Qwen-2.5-7B, and analyze whether light continual pretraining (CPT) on parallel data provides a useful foundation for post-training. No single method dominates: SFT maximises in-distribution accuracy but not consistency, DCO yields the strongest consistency gains but fails to transfer to free-form generation, and our GRPO variant achieves the strongest transfer to free-form recall and unseen languages on the multilingual base model. CPT mildly aids monolingual models but harms multilingual ones. Mechanistic analyses suggest GRPO is associated with reduced language specialization, consistent with greater sharing of representations across languages. We release our code, models and dataset publicly.
\footnote{
\simpleicon{huggingface}\ \href{https://huggingface.co/datasets/jvonrad/PolyFact}{\texttt{jvonrad/PolyFact} \\
\faGithub\ \href{https://github.com/jvonrad/Lost-in-Mistranslation}{\texttt{jvonrad/Lost-in-Mistranslation}} \\}
} \end{abstract}


\section{Introduction}

Large language models (LLMs) trained predominantly on English data encode vast amounts of world knowledge, yet struggle to reliably access this knowledge in other languages, leading to \textit{cross-lingual factual inconsistency} \cite{wang2025lost, schut2025do}. This raises a key question: how can models be enabled to access their already present latent knowledge without the need for additional pretraining?

\begin{figure}[t]
    \centering
    \includegraphics[width=1.0\linewidth]{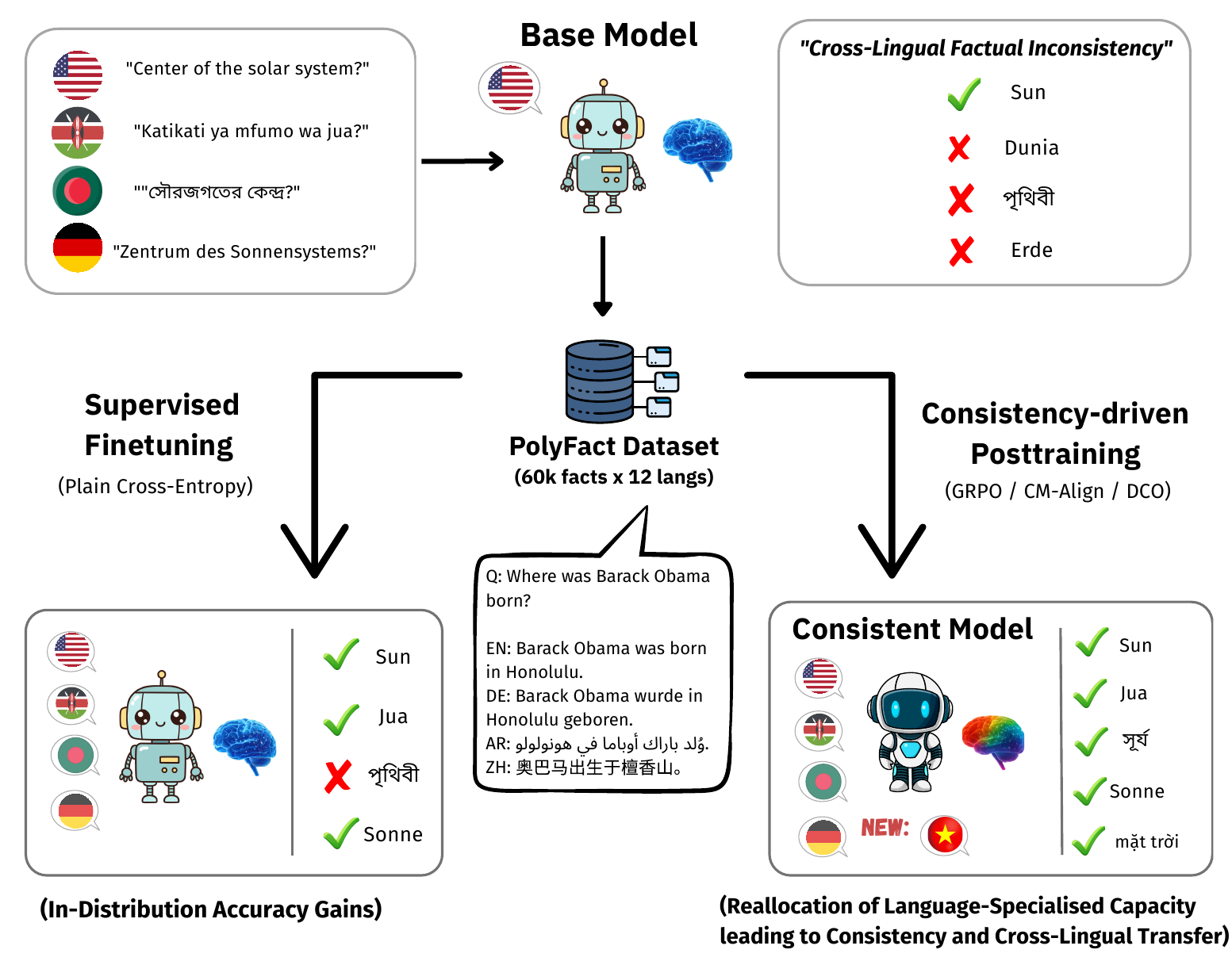}
   \caption{Incentivizing cross-lingual factual consistency through post-training on our multilingual question-answering dataset, \textsc{PolyFact}. Consistency-driven posttraining promotes shared internal representations that yield consistent factual predictions across languages, whereas SFT primarily leads to surface-level memorization.}
\label{fig:main-figure}
\end{figure}

Recent work suggests that multilingual models may rely on shared internal representations \cite{schut2025do, wendler-etal-2024-llamas}, where reasoning is performed in a shared latent space before being translated into the target language. In line with this perspective, prior work has shown that cross-lingual factual inconsistency often does not stem from missing knowledge, but rather emerges during the \emph{language transition phase} \cite{wang2025lost, gekhman2025insideout, lu-etal-2025-paths, liu2025entity}. Specifically, models may correctly retrieve the answer in intermediate layers, yet fail to map it reliably into the target language in later layers, leading to inconsistent or incorrect outputs across languages.

More recently, parallel data has been identified as a key driver of multilingual capabilities during pretraining \cite{qorib2025parallel, shao2026rolemixedlanguagedocumentsmultilingual, wang-etal-2025-investigating-scaling, qorib2025parallel, fu2024the, lin2025from, wu-etal-2024-far}. However, while continual pretraining (CPT) on parallel corpora improves translation fluency, it often fails to substantially improve performance on more demanding tasks such as multilingual factual recall \cite{shen2025unaligned}. This suggests that parallel data primarily improves the alignment of internal representations, while the model still struggles to reliably access knowledge encoded by the aligned representations through non-English language interfaces, resulting in inconsistent multilingual outputs.

Building on this insight, we hypothesize that multilingual factual recall in English-dominant LLMs can be improved without large-scale retraining, by separating \emph{representation alignment} from \emph{cross-lingual knowledge access}. 
Concretely, our contributions are as follows:
\begin{enumerate}[label=(\roman*), nosep]
\item We introduce \textsc{PolyFact}, a fully parallel multilingual factual QA dataset of 58{,}807 Wikidata-grounded facts across 12 typologically diverse languages.
\item Consistency-driven GRPO with cross-lingual reward pooling, which normalizes advantages jointly over all 12 languages' rollouts, concentrating gradient on cross-lingually inconsistent facts, compared with SFT, CM-Align and DCO in a controlled setup.
\item Evidence that no single post-training method dominates: the base model's pretraining mix determines the right method, with our GRPO variant transferring best to free-form generation and unseen languages on multilingual models, and CPT helping monolingual but harming multilingual models.
\end{enumerate}


\section{Related Work}
\paragraph{Cross-Lingual Factual Consistency.}
LLMs often recall the same facts unevenly across languages \cite{jiang-etal-2020-x, fierro-sogaard-2022-factual}. \citet{qi-etal-2023-cross} formalise this with RankC, which measures weighted top-$j$ candidate-set overlap and therefore captures agreement across the answer ranking rather than only top-1 predictions; we adopt it as our primary consistency metric. Mechanistic studies suggest that these gaps arise not necessarily from missing knowledge, but from failures either to map prompts into shared language-agnostic representations or to decode latent concepts into the correct target-language tokens \cite{wang2025lost, lu-etal-2025-paths, liu2025entity}. Existing remedies include inference-time subject injection, English pivoting, and activation patching \cite{bandarkar2026large, liu2025entity, lu-etal-2025-paths}, as well as English-anchored training methods such as CM-Align and DCO \cite{zhang-etal-2025-cm, liu2026posttraininglanguagemodelscrosslingual}. GRPO \cite{Guo_2025} has recently been applied to multilingual RAG and factual reasoning \cite{qi2026language, zhang-etal-2026-think}, but not to improving closed-book factual consistency across languages. We address this gap by analyzing GRPO with cross-lingual reward pooling.

\paragraph{Parallel Data.}
Extending English-centric models commonly relies on continual pretraining, which is computationally expensive and susceptible to catastrophic forgetting \cite{fujii2024continual, kuulmets2024teaching}. Although parallel data is a major source of multilingual capability during pretraining \cite{qorib2025parallel, lin2025from}, CPT on parallel corpora may improve translation without benefiting harder tasks such as cross-lingual factual recall \cite{shen2025unaligned}. This suggests that greater fluency does not necessarily translate into more consistent access to internal knowledge, a pattern also reflected in our results.

\section{Method}
\label{methods}

\begin{figure}[h]
\centering
\includegraphics[width=\columnwidth]{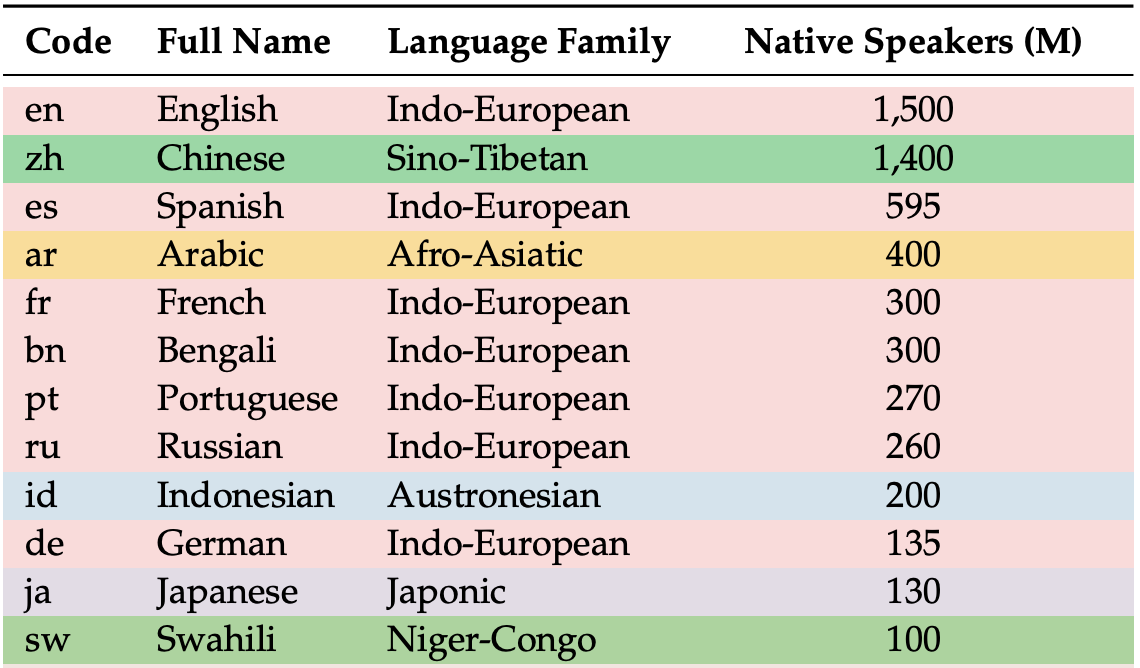}
\caption{Expanding language coverage from English-only capability to the 12 most widely spoken languages $(18.5\% \rightarrow 70\% $ of the global population).}
\label{fig:coverage}
\end{figure}

\paragraph{PolyFact Dataset.}
We construct \textsc{PolyFact}, a fully parallel multilingual multiple-choice
QA dataset for studying cross-lingual factual consistency. Starting from
Wikidata truthy triples,\footnote{\url{https://dumps.wikimedia.org/wikidatawiki/entities/latest-all.json.bz2}}
we curate 22 factual relations spanning geography, biography, creative works,
and organizational or cultural ties, and extract labels in twelve high- and
low-resource languages (Figure~\ref{fig:coverage}). For each fact, we sample
three type- and length-matched distractors from co-occurring objects of the
same property, then use round-robin balanced sampling to obtain 100{,}113
candidate facts, generating parallel MCQ bundles with Gemma-3-27B-IT
\citep{gemmateam2025gemma3}. The candidate corpus then passes through a
multi-stage quality pipeline: a web-grounded GPT-4o judge cross-checked
against human annotation (91\% agreement), Wikidata-grounded audits of every
answer label, mechanical filters for answer leakage and answer-distribution
balance, and an exhaustive LLM-assisted review of \emph{every} evaluation item
in every language, with confirmed defects repaired or removed. The released
corpus contains 58{,}807 facts across 14 relations (56{,}324 training, 444
validation, 2{,}039 test), and every evaluation item carries a per-language
verification flag. Further details are provided in Appendix~\ref{app:polyfact}.

\paragraph{Continual Pretraining.}
We continually pretrain on \textsc{TED2025} \cite{shen2025unaligned}, a multi-way parallel corpus covering the 12 languages in Figure~\ref{fig:coverage}. We retain talks containing at least two target languages and format each row as a multilingual block where available translations appear once, in randomized order, followed by an end-of-sequence token. Adjacent rows from the same talk are packed into $\sim$512-token chunks and truncated at 1024 tokens. To improve coverage for Swahili and Bengali, we augment TED2025 with Rogendo English--Swahili and AI4Bharat Samanantar English--Bengali sentence pairs. The final CPT corpus contains 325{,}134 packed chunks totalling 235.5M tokens (Table~\ref{tab:cpt-token-counts}).

\paragraph{Post-Training via GRPO.}
We apply a multilingual variant of GRPO \cite{shao2024deepseekmath} on the \textsc{PolyFact} dataset, where each training item is a single 
factual multiple-choice question available in parallel across all twelve languages. For every fact we 
sample $G=8$ grouped rollouts; each rollout consists of twelve \emph{independent} generations, one per language, 
produced from language-specific prompts that instruct the model to return the answer in the target language. For a rollout producing answers $\{\hat{y}_\ell\}_{\ell=1}^{L}$ across $L=12$ languages, the reward is computed as
\begin{align}
R &= \sum_{\ell=1}^{L} r_\ell \;+\; \mathbb{1}\!\left[\forall \ell,\; r_\ell = 1\right], \\[4pt]
r_\ell &=
\begin{cases}
+1 & \hat{y}_\ell \text{ correct option} \\
-0.5 & \hat{y}_\ell \text{ hallucination}
\\
\phantom{+}0 & \hat{y}_\ell \text{ incorrect option}
\end{cases}
\end{align}
where an answer is considered correct if $\hat{y}_\ell$ matches the gold answer $y^\star_\ell$, and invalid if it does not match any option in $\mathcal{O}_\ell$. The final term adds a bonus of $+1$ if all languages are answered correctly, encouraging cross-lingual consistency. Unlike standard GRPO, which normalizes advantages within one language's rollouts, we pool each fact's $G$ multilingual rollouts into a single normalization group, so the advantage is non-zero only where rollouts disagree across languages. This acts as an implicit curriculum concentrating gradient on the model's cross-lingually inconsistent facts and, together with the all-correct bonus, optimizes for correct consistency rather than per-language correctness alone.

\paragraph{Mechanistic Interpretability -- LAPE}
\label{sec:lape}
To identify language-specific MLP neurons, we apply LAPE \cite{LAPE}, treating each intermediate feed-forward dimension as active when its post-activation value is positive. Using parallel \texttt{TED2025} text across the 12 languages \cite{shen2025unaligned}, we compute each neuron's activation frequency per language, normalize across languages, and calculate its Shannon entropy:

\[
LAPE_{i,j} = -\sum_k p'*{i,j,k}\log p'*{i,j,k}.
\]

Lower entropy indicates greater language specialization. Following \citet{LAPE}, we select the bottom 1\% of neurons, assign each to the language with its highest activation frequency, and compare their distribution across layers and languages.

\section{Experimental Setup}

\paragraph{Models.}
We evaluate our approach on OLMo-2-1124-7B \cite{olmo2_2024} and Qwen-2.5-7B \cite{qwen2025qwen25technicalreport}, two 7B-scale decoder-only language models with different pretraining distributions and multilingual capabilities. OLMo-2-1124-7B serves as an English-dominant base model (documented in its open-source pretraining corpus) with a large English--non-English performance gap, while Qwen-2.5-7B provides a stronger multilingual comparison point.

\paragraph{Training Pipeline.}
To disentangle the effects of continual pretraining (CPT) from those of SFT and GPRO post-training, we study six model variants: (i) the base 7B models, (ii) a CPT model obtained by adapting the base model on 235.5M tokens of balanced parallel data, (iii-iv) SFT and GRPO models trained directly from the base model, and (v-vi) SFT and GRPO models initialized from the CPT checkpoint. This setup allows us to isolate the contribution of light multilingual continual pretraining and consistency-driven post-training, as well as their interaction. To provide comparison with other consistency post-training methods, we additionally train and evaluate two of the most prevalent approaches: CM-Align \cite{zhang-etal-2025-cm} and DCO \cite{liu2026posttraininglanguagemodelscrosslingual}.

\paragraph{Training Setup and Evaluation.}
Experiments were conducted on A100 GPUs. All post-training techniques use LoRA finetuning \cite{hu2022lora} with rank $r=64$ and $\alpha=128$ to avoid catastrophical forgetting and enabling rigorous comparison between results. Continual pretraining uses 235.5M tokens from the balanced TED2025 parallel corpus. Detailed CPT training pipeline and hyperparameters are listed in Appendix \ref{app:hyperparams}.

For post-training, we train SFT, GRPO, DCO and CM-Align on \textsc{PolyFact}, either directly from the base model or from the CPT checkpoint. GRPO uses grouped rollouts with group size $G=8$ over $L=12$ languages and optimizes a reward that favors correct answers, penalizes hallucinated outputs, and adds a bonus when all languages are answered correctly. Full GRPO and SFT hyperparameters are reported in Appendix Tables~\ref{tab:grpo-hparams} and~\ref{tab:sft-hparams}.

We evaluate multilingual factual recall and consistency on \textsc{PolyFact}, KLAR, BMLAMA, and Global-MMLU \cite{wang2025lost, qi-etal-2023-cross, singh-etal-2025-global}. In addition to per-language accuracy, we report RankC \cite{qi-etal-2023-cross} and total consistency (here: proportion of parallel facts answered correctly in all languages). Evaluations use LightEval \cite{lighteval} and lm-evaluation-harness \cite{eval-harness}, with vLLM \cite{kwon2023efficient} as the serving backend. We decontaminate all benchmarks by removing overlapping items the relation, entity and exact-fact levels and report performance on non-overlapping subsets. More details in appendix \ref{app:benchmarks}. Finally, we use LAPE \cite{LAPE} as correlational diagnostics of differences in the layer-wise distribution and language assignment of MLP neurons.


\section{Results}
We first present the main performance results across datasets, before analysing cross-lingual transfer and mechanistic interpretability results to understand the roots of improved cross-lingual factual consistency.

\begin{table*}[th!]
\centering
\footnotesize
\setlength{\tabcolsep}{2.8pt}
\renewcommand{\arraystretch}{0.95}
\begin{tabular*}{\textwidth}{@{\extracolsep{\fill}}lcccccccc|ccccc}
\toprule
 & \multicolumn{8}{c}{\textbf{Accuracy}} & \multicolumn{5}{c}{\textbf{Consistency}} \\
\cmidrule(lr){2-9}\cmidrule(lr){10-14}
 & \multicolumn{2}{c}{\textsc{PolyFact}} & \multicolumn{2}{c}{BMLAMA} & \multicolumn{2}{c}{G-MMLU} & \multicolumn{2}{c}{KLAR$^{\star}$} & \multicolumn{2}{c}{\textsc{PolyFact}} & BMLAMA & \multicolumn{2}{c}{G-MMLU} \\
\cmidrule(lr){2-3}\cmidrule(lr){4-5}\cmidrule(lr){6-7}\cmidrule(lr){8-9}\cmidrule(lr){10-11}\cmidrule(lr){12-12}\cmidrule(lr){13-14}
Method & High & Low & High & Low & High & Low & Train. & OOD & TotC & RankC & RankC & TotC & RankC \\
\midrule
\multicolumn{14}{c}{\textit{OLMo-2-1124-7B (monolingual)}} \\
\midrule
Baseline   & 46.7 & 37.5 & 18.4 & 15.4 & 47.9 & 35.2 & 24.6 & 13.3 & 1.7\,/\,1.5 & 57.3 & 37.0 & 3.8\,/\,1.0 & 55.0 \\
\midrule
SFT        & \cellcolor{PosGreen!34}\textbf{+4.1} & \cellcolor{PosGreen!34}\textbf{+3.7} & +0.2 & -0.1 & +0.1 & \cellcolor{NegRed!25}-1.1 & \cellcolor{PosGreen!4}+1.3 & -0.3 & +0.1\,/\,-0.4 & \cellcolor{NegRed!4}-0.6 & \cellcolor{PosGreen!8}+0.4 & \cellcolor{NegRed!34}-1.8\,/\,+1.0 & \cellcolor{PosGreen!24}+2.3 \\
DCO        & \cellcolor{PosGreen!7}+0.9 & \cellcolor{PosGreen!19}+2.0 & \cellcolor{PosGreen!31}+2.6 & \cellcolor{PosGreen!34}\textbf{+2.1} & \cellcolor{PosGreen!34}\textbf{+0.7} & \cellcolor{NegRed!13}-0.6 & \cellcolor{PosGreen!18}+6.3 & \cellcolor{PosGreen!15}+4.0 & \cellcolor{PosGreen!34}\textbf{+4.7\,/\,+2.6} & \cellcolor{PosGreen!34}\textbf{+5.9} & +0.2 & +0.2\,/\,+0.0 & \cellcolor{PosGreen!4}+0.4 \\
CM-Align   & \cellcolor{PosGreen!26}+3.1 & \cellcolor{PosGreen!10}+1.1 & \cellcolor{NegRed!8}-0.7 & \cellcolor{PosGreen!18}+1.1 & \cellcolor{NegRed!20}-0.4 & \cellcolor{NegRed!34}-1.5 & \cellcolor{PosGreen!34}\textbf{+11.8} & \cellcolor{PosGreen!34}\textbf{+8.9} & \cellcolor{PosGreen!19}+2.6\,/\,+0.1 & \cellcolor{PosGreen!9}+1.6 & \cellcolor{PosGreen!12}\textbf{+0.5} & \cellcolor{NegRed!34}-1.8\,/\,+0.2 & \cellcolor{PosGreen!12}+1.1 \\
GRPO       & \cellcolor{PosGreen!23}+2.7 & \cellcolor{PosGreen!13}+1.4 & \cellcolor{PosGreen!34}\textbf{+2.8} & \cellcolor{PosGreen!11}+0.7 & \cellcolor{PosGreen!28}+0.6 & \textbf{-0.1} & \cellcolor{PosGreen!24}+8.4 & \cellcolor{PosGreen!23}+5.9 & +0.2\,/\,-0.1 & -0.1 & \cellcolor{NegRed!34}-1.4 & \textbf{+0.2\,/\,+1.5} & \cellcolor{PosGreen!34}\textbf{+3.3} \\

\midrule
\multicolumn{14}{c}{\textit{Qwen-2.5-7B (multilingual)}} \\
\midrule
Baseline   & 54.2 & 42.3 & 26.5 & 24.7 & 68.6 & 50.1 & 47.7 & 35.8 & 5.3\,/\,1.7 & 62.4 & 43.6 & 13.8\,/\,1.2 & 69.8 \\
\midrule
SFT        & \cellcolor{PosGreen!34}\textbf{+6.9} & \cellcolor{PosGreen!34}\textbf{+5.8} & \cellcolor{PosGreen!10}+0.6 & \cellcolor{PosGreen!34}+1.8 & \cellcolor{NegRed!23}-1.3 & \cellcolor{NegRed!31}-2.9 & \cellcolor{PosGreen!10}+2.8 & \cellcolor{PosGreen!17}+4.1 & \cellcolor{PosGreen!18}+3.4\,/\,-0.2 & \cellcolor{PosGreen!19}+3.3 & \cellcolor{PosGreen!34}\textbf{+0.6} & \cellcolor{NegRed!34}-5.0\,/\,+0.0 & \cellcolor{NegRed!34}-1.8 \\
DCO        & \cellcolor{PosGreen!26}+5.2 & \cellcolor{PosGreen!34}+5.8 & \cellcolor{PosGreen!34}\textbf{+2.0} & \cellcolor{PosGreen!34}\textbf{+1.8} & \cellcolor{NegRed!6}\textbf{-0.4} & \cellcolor{NegRed!5}\textbf{-0.5} & -0.6 & +0.1 & \cellcolor{PosGreen!34}\textbf{+6.5\,/\,+1.3} & \cellcolor{PosGreen!34}\textbf{+5.9} & \cellcolor{PosGreen!28}+0.5 & -0.5\,/\,+0.5 & -0.1 \\
CM-Align   & \cellcolor{PosGreen!14}+2.9 & \cellcolor{PosGreen!11}+1.8 & +0.3 & -0.2 & \cellcolor{NegRed!34}-2.0 & \cellcolor{NegRed!34}-3.2 & \cellcolor{PosGreen!20}+5.6 & \cellcolor{PosGreen!13}+3.2 & \cellcolor{PosGreen!9}+1.6\,/\,+0.2 & \cellcolor{PosGreen!7}+1.1 & -0.1 & \textbf{+0.2\,/\,+2.8} & \cellcolor{PosGreen!30}\textbf{+1.6} \\
GRPO       & \cellcolor{PosGreen!25}+5.2 & \cellcolor{PosGreen!20}+3.4 & +0.1 & \cellcolor{NegRed!8}-0.4 & \cellcolor{NegRed!11}-0.6 & \cellcolor{NegRed!22}-2.1 & \cellcolor{PosGreen!34}\textbf{+9.4} & \cellcolor{PosGreen!34}\textbf{+8.2} & \cellcolor{PosGreen!12}+2.4\,/\,+0.0 & \cellcolor{PosGreen!9}+1.6 & -0.3 & \cellcolor{NegRed!9}-1.3\,/\,+0.2 & \cellcolor{NegRed!8}-0.4 \\
\bottomrule
\end{tabular*}
\caption{Comparison of post-training methods (baseline in absolute values, methods relative to them). Evaluated on \textsc{PolyFact} test (in-distribution factual recall), BMLAMA-53 (out-of-distribution factual recall), Global-MMLU-Lite (harder knowledge task) and KLAR (free-form generation$^{\star}$ factual recall) across seven trained languages (Train.) and 10 unseen languages (OOD). High-resource = \{en, de, pt, ar, es, ru, fr, ja, zh\}; low-resource = \{id, bn, sw\}. TotC = ratio of facts (correct / wrong) consistent across all languages.}
\label{tab:main_results}
\end{table*}

\subsection{Comparison of Post-Training Methods}
\label{sec:posttraining}

Table~\ref{tab:main_results} shows that no single post-training method dominates. The methods trade off between in-distribution accuracy, cross-lingual consistency, and free-form transfer, with the trade-offs strongly modulated by whether the base model was pretrained predominantly on English (OLMo-2) or multilingually (Qwen-2.5).

\paragraph{SFT: format gains without knowledge access.}
SFT delivers the largest in-distribution \textsc{PolyFact} accuracy gains (OLMo: $+4.1$/$+3.7$pp; Qwen: $+6.9$/$+5.8$pp on high-/low-resource languages), but little beyond: BMLAMA gains are marginal, Global-MMLU degrades on Qwen ($-1.3$/$-2.9$pp), and on OLMo consistency stays flat despite the accuracy jump, the signature of per-language memorisation of the multiple-choice format rather than consolidated knowledge. On the multilingual Qwen, the same supervision does lift consistency ($+3.4$pp TotC, $+3.3$pp RankC), suggesting per-language supervision propagates across languages when representations are already partially shared, whereas on English-centric OLMo it lands in language-specific circuitry.

\paragraph{DCO: the consistency specialist.}
DCO achieves the largest in-distribution consistency gains on both models (\textsc{PolyFact} TotC $+4.7$/$+6.5$pp, RankC $+5.9$pp on both), the best BMLAMA accuracy on OLMo, and the mildest Global-MMLU footprint on Qwen, matching its design: aligning output distributions across parallel prompts directly optimises what RankC and TotC measure. Its limitation is free-form generation: moderate transfer to KLAR on OLMo ($+6.3$/$+4.0$pp), near-complete failure on Qwen ($-0.6$/$+0.1$pp). We hypothesise that distribution alignment mainly redistributes probability mass among already-accessible answers, leaving little to move on an already-aligned multilingual base and nothing that helps without candidates.

\paragraph{CM-Align: English pivoting suits English-centric models.}
CM-Align shows the mirror-image profile of DCO: moderate consistency gains but strong free-form transfer, dominating KLAR on OLMo ($+11.8$/$+8.9$pp, the largest transfer result in the table). Its EN-pivot construction anchors every language to the model's English behaviour, a strong teacher signal precisely when English is the model's dominant language, but a weaker, partially redundant one on the multilingual Qwen ($+5.6$/$+3.2$pp). It also consistently taxes broader knowledge, degrading Global-MMLU on both models (up to $-2.0$/$-3.2$pp on Qwen).

\paragraph{Consistency-driven GRPO: pivot-free access gains.}
Our GRPO variant achieves the strongest free-form transfer on Qwen (KLAR $+9.4$/$+8.2$pp, the best OOD-language gain of any method) and solid transfer on OLMo ($+8.4$/$+5.9$pp), without CM-Align's Global-MMLU cost. Its cross-lingual reward pooling concentrates gradient on facts answered inconsistently across languages, without privileging a pivot language: this yields the only notable out-of-distribution consistency gain in the table (OLMo Global-MMLU RankC $+3.3$pp, TotC $+0.2$/$+1.5$pp) at near-flat accuracy, i.e.\ GRPO makes answers more mutually consistent even where it cannot make them more accurate. Its smaller \textsc{PolyFact} consistency gains on Qwen ($+2.4$pp TotC) follow from the same mechanism: the advantage signal is non-zero only where rollouts disagree, and a multilingual base offers less inconsistency to exploit.

\paragraph{Summary.}
Token-level supervision improves the trained format, while distribution alignment improves consistency but not candidate-free access. Generation-based objectives improve free-form access with the best teacher signal depending on the base model: English pivoting for an English-centric model, cross-lingual reward pooling for a multilingual one. The pretraining mix is thus a first-order determinant of which post-training method is appropriate.

\begin{table*}[th!]
\centering
\footnotesize
\setlength{\tabcolsep}{2.8pt}
\renewcommand{\arraystretch}{0.95}
\begin{tabular*}{\textwidth}{@{\extracolsep{\fill}}lcccccccc|ccccc}
\toprule
 & \multicolumn{8}{c}{\textbf{Accuracy}} & \multicolumn{5}{c}{\textbf{Consistency}} \\
\cmidrule(lr){2-9}\cmidrule(lr){10-14}
 & \multicolumn{2}{c}{\textsc{PolyFact}} & \multicolumn{2}{c}{BMLAMA} & \multicolumn{2}{c}{G-MMLU} & \multicolumn{2}{c}{KLAR$^{\star}$} & \multicolumn{2}{c}{\textsc{PolyFact}} & BMLAMA & \multicolumn{2}{c}{G-MMLU} \\
\cmidrule(lr){2-3}\cmidrule(lr){4-5}\cmidrule(lr){6-7}\cmidrule(lr){8-9}\cmidrule(lr){10-11}\cmidrule(lr){12-12}\cmidrule(lr){13-14}
 & High & Low & High & Low & High & Low & Train. & OOD & TotC & RankC & RankC & TotC & RankC \\
\midrule
\multicolumn{14}{c}{\textit{OLMo-2-1124-7B (monolingual)}} \\
\midrule
CPT      & -0.2 & \cellcolor{PosGreen!5}+0.3 & \cellcolor{NegRed!10}-0.4 & \cellcolor{NegRed!6}-0.3 & \cellcolor{NegRed!34}-1.8 & \cellcolor{NegRed!31}-1.6 & \cellcolor{NegRed!19}-7.5 & \cellcolor{NegRed!17}-5.0 & \cellcolor{PosGreen!21}+1.1\,/\,-0.0 & \cellcolor{PosGreen!25}+1.3 & \cellcolor{PosGreen!32}+1.5 & \cellcolor{NegRed!34}-1.5\,/\,+0.5 & \cellcolor{NegRed!29}-0.5 \\
\;+\,SFT      & \cellcolor{PosGreen!34}+1.4 & \cellcolor{PosGreen!34}+2.3 & +0.0 & \cellcolor{PosGreen!11}+0.6 & \cellcolor{NegRed!14}-0.7 & \cellcolor{NegRed!15}-0.8 & \cellcolor{NegRed!26}-10.3 & \cellcolor{NegRed!18}-5.2 & \cellcolor{PosGreen!20}+1.0\,/\,+0.2 & \cellcolor{PosGreen!34}+1.7 & \cellcolor{PosGreen!28}+1.3 & \cellcolor{PosGreen!11}+0.5\,/\,-0.2 & -0.2 \\
\;+\,DCO     & +0.3 & \cellcolor{PosGreen!10}+0.7 & \cellcolor{PosGreen!13}+0.5 & \cellcolor{NegRed!13}-0.7 & \cellcolor{NegRed!12}-0.6 & \cellcolor{PosGreen!34}+1.7 & \cellcolor{NegRed!34}-13.4 & \cellcolor{NegRed!34}-9.7 & \cellcolor{PosGreen!34}+1.7\,/\,+0.6 & \cellcolor{PosGreen!18}+0.9 & \cellcolor{PosGreen!34}+1.6 & \cellcolor{NegRed!28}-1.2\,/\,+0.8 & \cellcolor{NegRed!34}-0.5 \\
\;+\,GRPO    & \cellcolor{PosGreen!15}+0.6 & \cellcolor{PosGreen!24}+1.6 & \cellcolor{PosGreen!34}+1.4 & \cellcolor{PosGreen!34}+1.9 & \cellcolor{NegRed!30}-1.6 & \cellcolor{PosGreen!8}+0.4 & \cellcolor{NegRed!8}-3.1 & \cellcolor{NegRed!6}-1.6 & \cellcolor{PosGreen!17}+0.8\,/\,-0.2 & \cellcolor{PosGreen!10}+0.5 & \cellcolor{NegRed!14}-0.7 & \cellcolor{NegRed!23}-1.0\,/\,+0.0 & -0.2 \\
\midrule
\multicolumn{14}{c}{\textit{Qwen-2.5-7B (multilingual)}} \\
\midrule
CPT       & \cellcolor{NegRed!25}-5.2 & \cellcolor{NegRed!34}-2.7 & \cellcolor{NegRed!27}-1.2 & -0.1 & \cellcolor{NegRed!33}-2.9 & \cellcolor{NegRed!34}-3.1 & \cellcolor{NegRed!24}-8.8 & \cellcolor{NegRed!24}-12.0 & \cellcolor{NegRed!14}-1.4\,/\,+0.3 & \cellcolor{NegRed!12}-1.2 & +0.2 & \cellcolor{NegRed!5}-0.5\,/\,+0.8 & \cellcolor{NegRed!34}-1.5 \\
\;+\,SFT     & \cellcolor{NegRed!21}-4.3 & \cellcolor{PosGreen!32}+2.5 & \cellcolor{NegRed!34}-1.5 & \cellcolor{NegRed!21}-0.9 & \cellcolor{NegRed!27}-2.4 & +0.1 & \cellcolor{NegRed!34}-12.5 & \cellcolor{NegRed!34}-17.1 & \cellcolor{NegRed!34}-3.3\,/\,+0.4 & \cellcolor{NegRed!21}-2.1 & +0.1 & \cellcolor{PosGreen!34}+3.2\,/\,+0.8 & \cellcolor{PosGreen!19}+0.9 \\
\;+\,DCO      & \cellcolor{NegRed!13}-2.8 & \cellcolor{NegRed!6}-0.5 & \cellcolor{NegRed!24}-1.0 & \cellcolor{PosGreen!19}+0.8 & \cellcolor{NegRed!34}-2.9 & \cellcolor{NegRed!20}-1.8 & \cellcolor{NegRed!22}-8.3 & \cellcolor{NegRed!19}-9.7 & \cellcolor{PosGreen!16}+1.6\,/\,+1.1 & \cellcolor{PosGreen!4}+0.4 & +0.2 & \cellcolor{PosGreen!13}+1.2\,/\,+0.8 & \cellcolor{NegRed!17}-0.8 \\
\;+\,GRPO    & \cellcolor{NegRed!34}-7.1 & \cellcolor{NegRed!12}-1.0 & \cellcolor{NegRed!14}-0.6 & \cellcolor{PosGreen!34}+1.5 & \cellcolor{NegRed!25}-2.2 & \cellcolor{PosGreen!9}+0.8 & \cellcolor{NegRed!30}-11.1 & \cellcolor{NegRed!26}-13.2 & \cellcolor{NegRed!30}-2.9\,/\,-0.2 & \cellcolor{NegRed!34}-3.4 & \cellcolor{PosGreen!34}+0.5 & \cellcolor{NegRed!8}-0.8\,/\,+0.2 & \cellcolor{NegRed!17}-0.8 \\
\bottomrule
\end{tabular*}
\caption{Effect of continued pre-training. Each row is the change from applying the method on top of the CPT checkpoint rather than the base model, i.e.\ CPT\,+\,X minus X; the first row of each block is CPT alone against the base model. All X-references are the same models as in Table~\ref{tab:main_results}. Shading and columns as in Table~\ref{tab:main_results}.}
\label{tab:cpt_effect}
\end{table*}

\subsection{Effect of Continued Pretraining}
\label{sec:cpt}

Table~\ref{tab:cpt_effect} shows that light parallel-data CPT is at best neutral and often harmful, splitting along the same monolingual/multilingual axis. On the English-centric OLMo, CPT alone leaves closed-form accuracy unchanged and buys small consistency gains (\textsc{PolyFact} RankC $+1.3$pp, BMLAMA RankC $+1.5$pp): parallel text mildly tightens cross-lingual alignment the monolingual pretraining never provided, and stacked under post-training it adds modest closed-form accuracy (e.g., CPT$+$SFT $+1.4$/$+2.3$pp on \textsc{PolyFact}). On the multilingual Qwen, CPT is damaging nearly everywhere: alone it costs $-5.2$/$-2.7$pp on \textsc{PolyFact} and $-8.8$/$-12.0$pp on KLAR, and it degrades most downstream combinations, most severely CPT$+$GRPO ($-7.1$pp high-resource, $-3.4$pp RankC). A model whose representations are already aligned has little to gain from additional parallel signal, and the distribution shift only perturbs it.

The most consistent effect is shared by both models: a severe loss of free-form recall. Every CPT$+$X variant underperforms its non-CPT counterpart on KLAR, with drops up to $-13.4$/$-9.7$pp (OLMo CPT$+$DCO) and $-12.5$/$-17.1$pp (Qwen CPT$+$SFT); CPT$+$GRPO on OLMo is least affected ($-3.1$/$-1.6$pp), consistent with on-policy training partially repairing generation behaviour. That closed-form metrics stay flat or improve while free-form recall collapses points to an interface-level disruption rather than knowledge loss: CPT on concatenated parallel text perturbs generation behaviour, which candidate ranking masks but open-ended prompting exposes. In short, parallel-data CPT mildly aids a monolingual model's closed-form cross-lingual alignment, contributes little factual knowledge, damages free-form access, and disrupts multilingual base models outright.

\subsection{Language Transition}

\begin{figure}[t!]
    \centering
    \begin{subfigure}[t]{\linewidth}
        \centering
        \includegraphics[width=\linewidth]{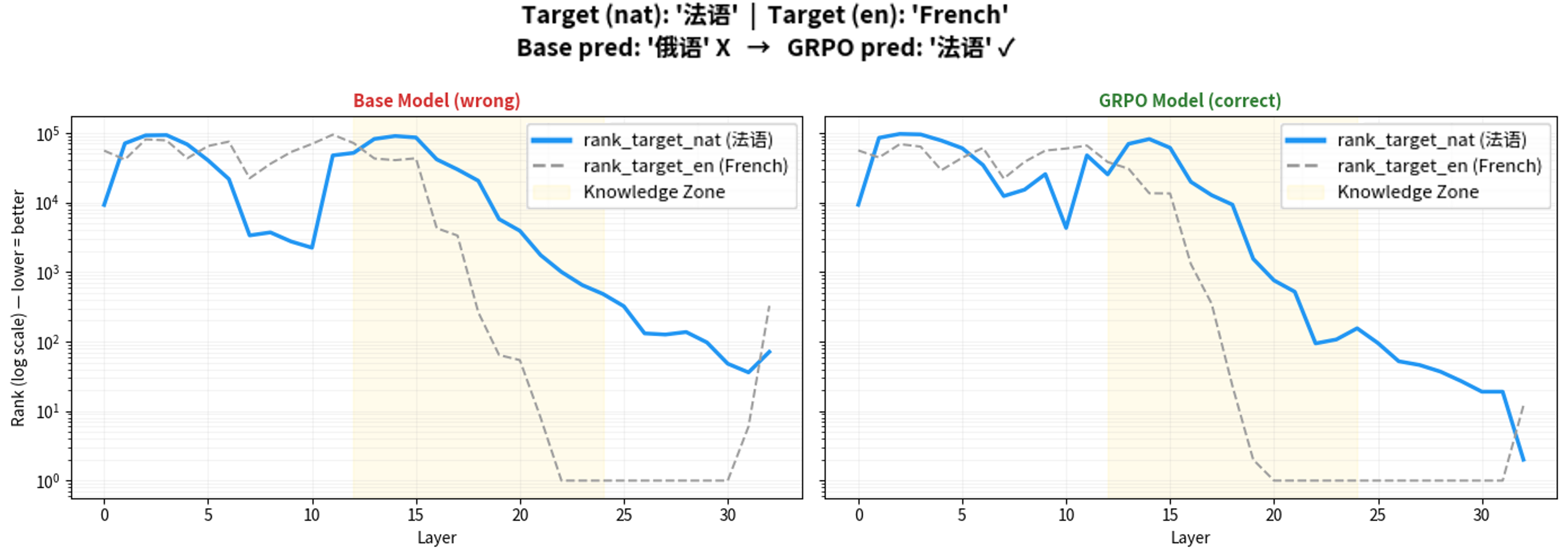}
        \caption{[ZH] Official language: "French Polynesia"}
        \label{fig:zh_plot_flat}
    \end{subfigure}
    
    \vspace{1pt}
    
    \begin{subfigure}[t]{\linewidth}
        \centering
        \includegraphics[width=\linewidth]{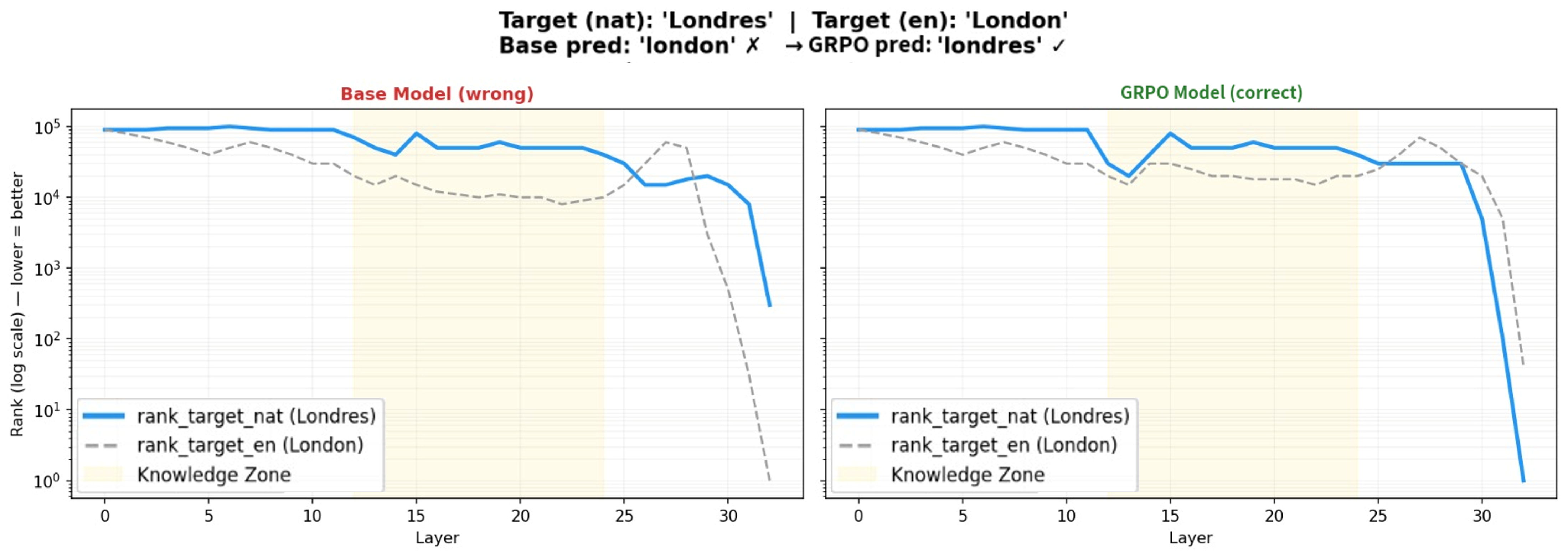}
        \caption{[ES] Place of birth: "Mark Strong"}
        \label{fig:mark_strong}
    \end{subfigure}
    
    \caption{Layer-rank analysis of Base (left) and GRPO finetuned OLMO-2 (right) models, showing rank of correct English and target language tokens. 
    (a) Example in non-Latin script language. 
    (b) Example in Latin-script language.}
    \label{fig:language_transition}
\end{figure}

To understand how GRPO improves cross-lingual behaviour, we analyse the layer-wise ranking of the correct answer in both the target language and English (Figures~\ref{fig:mark_strong} and~\ref{fig:zh_plot_flat}). These plots show where the model retrieves the correct fact and how it transitions into the target language.

In the base model, a consistent failure mode emerges: the correct answer is retrieved in English (grey dashed line ranks highly in later layers), while the target-language form (blue line) remains poorly ranked. This indicates that the model accesses the correct knowledge but fails to convert it into the required language. In the Spanish example (Figure~\ref{fig:mark_strong}), the model retrieves \textit{London} but fails to produce \textit{Londres}; similarly, in the Chinese example (Figure~\ref{fig:zh_plot_flat}), the concept is accessible but not properly transferred. This reflects a breakdown in the final stage of processing: the transition from a shared or English-centric representation to the target-language output, aligned with findings from \citet{wang2025lost}.

Consistency-driven GRPO fine-tuning mitigates this issue. In both examples, the target-language answer rises in early layers and overtakes the English form in later layers. This suggests that GRPO strengthens the pathway between language-agnostic knowledge representations and target-language decoding, enabling the model to directly produce the correct linguistic form rather than relying on an English intermediate. A more detailed analysis of failure modes during factual retrieval can be found in Appendix \ref{app:failure-modes}.

\subsection{Cross-lingual Transfer}

We evaluate cross-lingual transfer on KLAR, which covers seventeen languages:
the seven present during post-training (en, es, fr, ru, zh, ja, ar) and ten
held out from it (ca, el, fa, he, hu, ko, nl, tr, uk, vi). Figure~4(a) reports
the two aggregates and Figure~4(b) the per-language breakdown.

GRPO improves KLAR accuracy from 24.6 to 33.0 on trained languages and from
13.3 to 19.2 on held-out ones (+8.4 / +5.9). The held-out gain is 70\% of the
trained-language gain, and the improvement is broad rather than concentrated:
accuracy rises in 15 of 17 languages, the exceptions being Greek ($-2.1$) and
Hebrew ($-0.4$). Among the held-out languages, Catalan (+12.6), Dutch (+11.5)
and Vietnamese (+10.7) improve as much as the trained ones, so the gains are
not confined to the training distribution.

The gains are largest where the base model is weakest. Arabic rises from 7.9 to
20.7 (+162\% relative), Japanese from 8.1 to 16.8 (+107\%) and Chinese from 9.8
to 15.2 (+55\%), against +17\% for English, which starts at 62.9. Post-training
therefore compresses the spread between the model's strongest and weakest
languages rather than lifting all of them uniformly -- the behaviour our
cross-lingual consistency objective is intended to produce, measured here in
free-form generation rather than in candidate ranking.

SFT moves KLAR by +1.3 on trained languages and
$-0.3$ on held-out ones, and improves only 10 of 17 languages. Its largest
single change is English (+4.4). Supervision on parallel data in each language
independently is thus not sufficient to produce transfer: the gains we observe
require an objective that couples the languages during training.

\begin{figure}[t]
    \centering
    \begin{subfigure}[t]{\linewidth}
        \centering
        \includegraphics[width=0.9\linewidth]{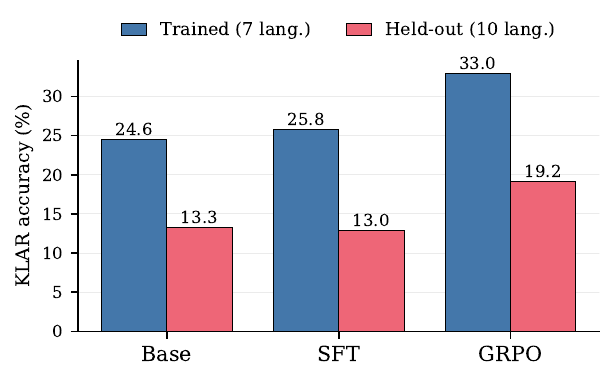}
        \caption{Accuracy on trained vs.\ held-out languages.}
        \label{fig:seen-unseen}
    \end{subfigure}
    
    \vspace{0pt}
    
    \begin{subfigure}[t]{\linewidth}
        \centering
        \includegraphics[width=\linewidth]{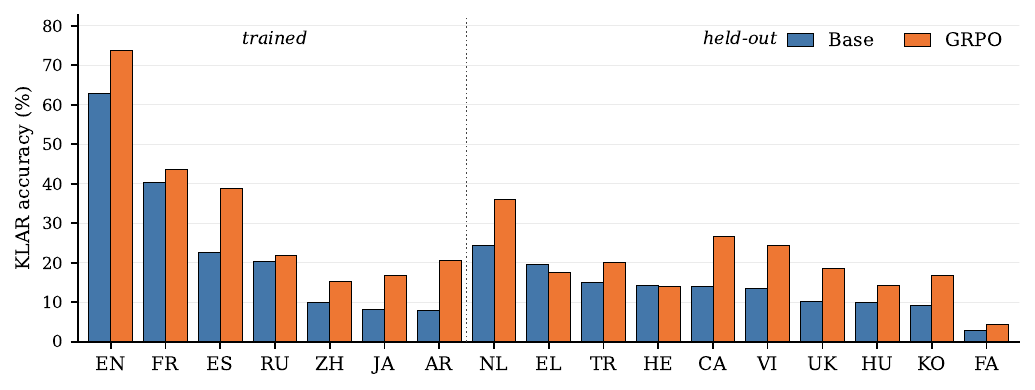}
        \caption{Per-language accuracy comparison.}
        \label{fig:klar-per-lan}
    \end{subfigure}
    
    \caption{OLMo-2-7B KLAR performance on trained and held-out languages across (a) models and (b) languages.}
    \label{fig:klar_combined}
\end{figure}

\begin{figure*}[th!]
\centering
\includegraphics[width=1\textwidth]{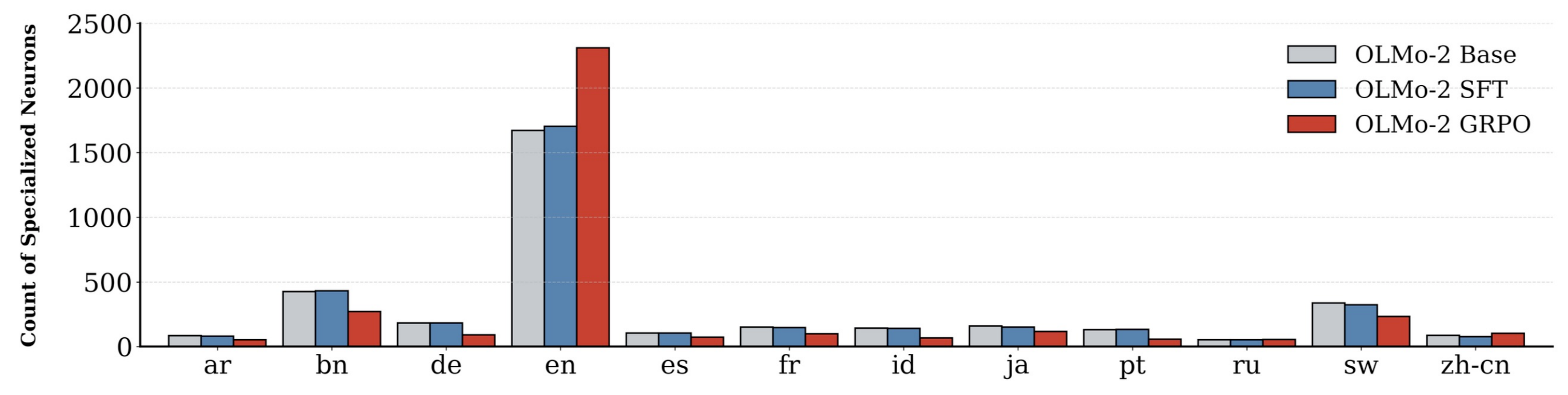}
\caption{Language-specific neurons across languages; GRPO increases English alignment while SFT does not change language-specialization considerably.}
\label{fig:lape_total}
\end{figure*}

\subsection{Language-Specific Neurons}

Following the LAPE methodology described in Section \ref{sec:lape}, we identify and evaluate the distribution of language-specific neurons in three configurations of OLMo-2-7B: the base model and \textsc{PolyFact} fine-tuned model variants using SFT and GRPO. 

\paragraph{Total neuron count shifts towards English-specificity.}
An analysis of the language-specific neuron distribution reveals a striking and somewhat counterintuitive effect. While the SFT model remains nearly identical to the Base model, GRPO exhibits a pronounced shift toward English-specific neurons. As shown in Figure~\ref{fig:lape_total}, GRPO increases English-specific neurons by 38.2\% (1671 to 2310), at the expense of other languages, with large reductions for Bengali (-36.1\%, 424 to 271) and Swahili (-31.2\%, 337 to 232), while Chinese is a notable exception (+18.6\%, 86 to 102).

Notably, this shift cannot be explained by improved English performance. The SFT model achieves stronger English \textsc{PolyFact} accuracy than GRPO, and GRPO itself performs slightly better than the baseline (Appendix \ref{app:per-lang-perf}, Table ~\ref{tab:wikifact_appendix}). This suggests the effect reflects a structural reorganisation rather than performance gains. One interpretation is that the RL objective ``squeezes'' the model’s behavioural space, encouraging reliance on the most stable representational backbone (here: English), consistent with the work of \citet{matsutani2025rl}.

\paragraph{GRPO delays linguistic specialisation in non-Latin scripts.}
To quantify the layer-wise distribution of language-specific neurons, we analyse the empirical cumulative distribution function (ECDF) across layers (Figure~\ref{fig:lape_ecdf} in Appendix \ref{app:lape_full}). We observe a clear ``late-discovery" effect unique to GRPO: while the base and SFT models show nearly identical distributions, the GRPO curve is systematically lower, indicating that specialised neurons are recruited later in the network. This is reflected in a negligible Kolmogorov-Smirnov distance for SFT ($D_{KS} = 0.005$) and a substantial shift for GRPO ($D_{KS} = 0.089$).

\begin{figure}[t]
    \centering
    \includegraphics[width=\linewidth]{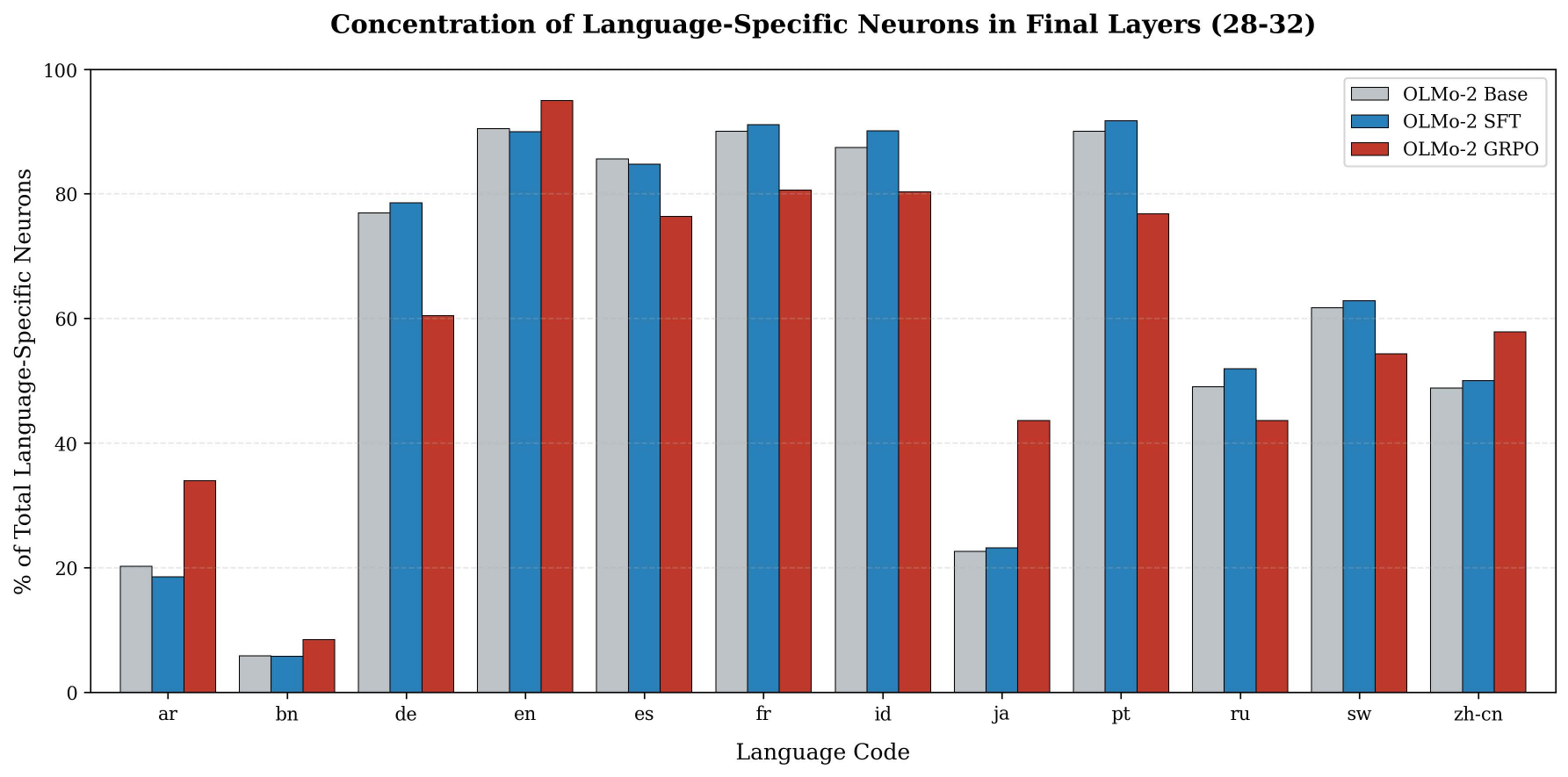}
    \caption{Concentration of specialised neurons in the final layers (28--32). GRPO exhibits a bifurcation: Latin scripts shift toward intermediate layers, while distinct scripts consolidate in the final layers leading script-dependent processing.}
    \label{fig:lape_script_concentration}
\end{figure}

The lower GRPO ECDF indicates that the selected low-entropy neurons tend to occur later in the network. The direction of this shift differs across scripts: neurons assigned to Latin-script languages occur relatively earlier, whereas those assigned to languages such as Arabic and Japanese are more concentrated in later layers (Figure~\ref{fig:lape_script_concentration}). Because LAPE measures relative activation specialisation and we do not intervene on these neurons, we interpret this as a change in the layer-wise localization of language-selective activity, not as evidence that the intervening layers become language-agnostic. A full breakdown is provided in Appendix~\ref{app:lape_full}.

\section{Conclusion}
We study cross-lingual factual recall and consistency in LLMs, contributing \textsc{PolyFact}, a fully parallel factual QA dataset of 60K Wikidata-grounded facts across 12 languages, and consistency-driven GRPO with cross-lingual reward pooling. Comparing against SFT, DCO and CM-Align on OLMo-2-1124-7B and Qwen-2.5-7B, no single method dominates: SFT improves the trained format, DCO yields the strongest consistency gains without improving candidate-free access, and our GRPO variant transfers best to free-form recall and held-out languages on multilingual base models, requiring no pivot language. Light parallel-data CPT mildly aids monolingual models as a foundation but harms multilingual ones and damages free-form recall across methods. Mechanistically, GRPO is associated with reduced language specialization in favour of shared cross-lingual computation, though these diagnostics remain correlational. Overall, cross-lingual factual inconsistency appears to be less a problem of missing knowledge than of unreliable access, and the right post-training remedy depends on the model's pretraining distribution.

\newpage

\section{Limitations}

Our work has several limitations. First, our experiments are conducted on two 7B model families, OLMo-2-7B and Qwen-2.5-7B. While these models cover different training setups and multilingual capabilities, it is not certain whether our findings generalise to larger models, smaller models, or substantially different architectures such as mixture-of-experts models.

Second, training on \textsc{PolyFact} does not fully generalise to more challenging benchmarks such as Global-MMLU, indicating limitations on tasks that require deeper reasoning beyond factual retrieval. Our method is therefore best understood as improving cross-lingual factual access rather than solving multilingual reasoning more broadly.

Finally, our work introduces potential risks. Since \textsc{PolyFact} is derived from Wikidata, it may inherit coverage biases, factual errors, or uneven representation across languages, regions, and entities. Moreover, improving cross-lingual factual recall may also make incorrect or biased factual associations more consistently expressed across languages if such information is present in the underlying model or dataset. We mitigate these risks through relation filtering, automatic and manual quality checks, and by releasing the dataset and code to support transparency, auditing, and future corrections.

\paragraph{LLMs Usage.} Through the paper, we use LLMs to assist with grammar checking and minor rephrasing for clarity. LLMs did not contribute to the conceptual design of the study, or core writing of the paper.

\bibliography{references}      

\newpage

\appendix

\section{Benchmark details}
\label{app:benchmarks}

We evaluate across three benchmarks targeting complementary aspects of
multilingual factual recall: factual recall in the training-task format
(\textsc{PolyFact}), free-form factual recall (KLAR-CLC) and broader
multilingual knowledge and reasoning (Global-MMLU). All inference uses bf16 precision.

\paragraph{PolyFact.}
We evaluate on the 2{,}500-fact test split of \textsc{PolyFact} using
a custom evaluator (\texttt{evaluate/evaluate\_consistency.py}) that
scores each of the four MCQ options by its length-normalised
conditional log-likelihood under the model, conditioning on a
language-specific prompt that wraps the question and instructs the
model to answer in the target language. The option with the highest
per-token average log-probability is selected. We report per-language
accuracy across all twelve target languages (en, de, es, fr, pt, id,
ru, ar, zh, ja, sw, bn). Because facts are fully parallel across
languages, per-language differences isolate language-interface
effects from underlying knowledge.

\paragraph{KLAR-CLC.}
For free-form cross-lingual factual recall we evaluate on KLAR-CLC
\cite{wang2025lost} using a custom evaluator
(\texttt{evaluate/evaluate\_klar.py}). The evaluator runs 3-shot
prompted greedy generation (max 10 new tokens) and compares the
generated answer to the gold using NFC-normalised, case-insensitive,
punctuation-stripped string matching, with the punctuation regex
covering ASCII, CJK, and typographic quotes/guillemets to handle
all 17 KLAR languages. KLAR is structured into 20 factual relations
(e.g.\ \textit{capital}, \textit{place of birth},
\textit{manufacturer}, \textit{occupation}). Unlike \textsc{PolyFact},
KLAR removes the candidate set, testing whether the model can
\emph{generate} the correct answer in the target language rather than
just discriminate among given options. We evaluate on six seen
languages (en, es, fr, ru, zh, ja) and eleven held-out languages
(ar, ca, el, fa, he, hu, ko, nl, tr, uk, vi) to measure cross-lingual
transfer.

\paragraph{Global-MMLU.}
For broader multilingual knowledge and reasoning we evaluate on
Global-MMLU \cite{singh-etal-2025-global} using
\texttt{lm-evaluation-harness} \cite{eval-harness} with vLLM
\cite{kwon2023efficient} as the backend (task identifiers
\texttt{global\_mmlu\_full\_\{lang\}}, \texttt{batch\_size=auto},
\texttt{gpu\_memory\_utilization=0.85}). Scoring is standard
log-likelihood over the four MCQ options. We report per-language
accuracy across the twelve target languages. Global-MMLU is
substantially harder than direct factual-recall benchmarks because
it additionally requires multi-step reasoning and domain knowledge,
making it a stricter test of generalisation beyond
training-distribution facts.

\paragraph{Decontimination.}
We remove all overlapping items from our evaluated benchmark sets to ensure faithful evaluation.
For KLAR, 8 of 20 relations overlap with \textsc{PolyFact}; 157 of 2{,}619 facts exactly overlap with its training set, corresponding to 13.0\% of the 1{,}207 examples from shared relations and 6.0\% of the complete benchmark. We report both denominators and separately evaluate facts and relations absent from \textsc{PolyFact}. For Global-MMLU, 2.3\% of items co-mention the subject and object of a \textsc{PolyFact} fact, while verbatim eight-gram question overlap is 0.05\%.


\paragraph{Inference backends.}
\texttt{lm-evaluation-harness} and LightEval use vLLM for throughput;
the \textsc{PolyFact} and KLAR custom evaluators use HuggingFace
Transformers directly because they need access to per-token
log-probabilities (\textsc{PolyFact}) and explicit batched greedy
decoding with custom string matching (KLAR), which are awkward to
express in the harnesses.

\section{Hyperparameters}
\label{app:hyperparams}

\paragraph{Continual Pretraining.}
We continually pretrain OLMo-2-1124-7B \cite{olmo2_2024} and Qwen-2.5-7B \cite{qwen2025qwen25technicalreport} on the balanced TED2025 dataset using the 
hyperparameters in Table~\ref{tab:cpt-hyperparams}. A detailed per-language of the TED2025 subset used for CPT is listed in Table \ref{tab:cpt-token-counts}.

\begin{table}[t]
\centering
\scriptsize
\setlength{\tabcolsep}{3pt}
\resizebox{\columnwidth}{!}{%
\begin{tabular}{ll}
\toprule
\multicolumn{2}{l}{\textit{Data}} \\
\midrule
Max sequence length         & 1024 tokens \\
Chunk packing target        & 512 tokens (talk-internal) \\
Min languages per row       & 2 \\
Min languages per talk      & 2 \\
Validation split            & 0.5\% (language-balanced) \\
\midrule
\multicolumn{2}{l}{\textit{Optimization}} \\
\midrule
Optimizer                   & AdamW (fused) \\
Learning rate               & $2\times10^{-5}$ \\
LR schedule                 & Cosine \\
Warmup steps                & 200 \\
Weight decay                & 0.01 \\
Precision                   & bf16 \\
Epochs                      & 1 \\
Per-device batch size       & 6 \\
Gradient accumulation steps & 8 \\
\bottomrule
\end{tabular}%
}
\caption{Continual pretraining hyperparameters.}
\label{tab:cpt-hyperparams}
\end{table}

\begin{table}[t]
\centering
\small
\begin{tabular}{lrr}
\toprule
Language & \# lang-lines & Tokens (M) \\
\midrule
English (en)    & 1{,}778{,}887 & 67.2 \\
Spanish (es)    &   656{,}921 & 19.5 \\
Arabic (ar)     &   615{,}808 & 15.3 \\
French (fr)     &   534{,}647 & 16.6 \\
Chinese (zh)    &   515{,}553 & 13.7 \\
Bengali (bn)    &   500{,}581 & 35.3 \\
Russian (ru)    &   469{,}035 & 16.0 \\
Japanese (ja)   &   356{,}325 &  4.9 \\
German (de)     &   297{,}306 &  9.8 \\
Indonesian (id) &   219{,}040 &  7.3 \\
Portuguese (pt) &   210{,}441 &  6.7 \\
Swahili (sw)    &   174{,}125 &  4.0 \\
\midrule
Total & 6{,}328{,}669 & 216.4 \\
\bottomrule
\end{tabular}
\caption{Per-language coverage in the TED 2025 CPT corpus tokenized with the Qwen-2.5
tokenizer. ``\# lang-lines'' counts how often each language appears as a section within a
packed chunk; a single TED chunk typically contains 5--12 language sections.
Token counts exclude the \texttt{\{lang\}:} format wrapper. The remaining
$\sim$19M tokens in the 235.5M total are end-of-sequence and other framing tokens.}
\label{tab:cpt-token-counts}
\end{table}

\paragraph{GRPO Post-Training.}
Starting from Olmo-2-1124-7B basemodel and CPT checkpoint, we run multilingual GRPO on \textsc{PolyFact} with the configuration in 
Table~\ref{tab:grpo-hparams}.

\begin{table}[h]
\centering
\footnotesize
\setlength{\tabcolsep}{4pt}
\renewcommand{\arraystretch}{1.05}
\begin{tabular}{@{}ll@{}}
\toprule
\textbf{Parameter} & \textbf{Value} \\
\midrule
Dataset & \textsc{PolyFact} \\
Languages ($L$) & 12 \\
Group size ($G$) & 8 \\
Max train facts & 40{,}000 \\
Epochs & 1 \\
Batch size & 1 fact \\
Learning rate & $1\!\times\!10^{-5}$ \\
LR schedule & cosine, 3\% warmup \\
Weight decay & 0.01 \\
Optimizer & AdamW \\
Grad clipping & 1.0 \\
Max prompt / completion & 512 / 48 \\
Temperature / top-$p$ & 0.7 / 0.95 \\
Repetition penalty & 1.5 \\
No-repeat $n$-gram & 3 \\
Correct reward & $+1$ \\
Hallucination penalty & $-0.5$ \\
All-correct bonus & $+1$ \\
KL coefficient ($\beta$) & 0.0 \\
LoRA $r$ / $\alpha$ / dropout & 64 / 128 / 0.05 \\
Precision & bf16 \\
Grad checkpointing & enabled \\
Eval frequency & every 500 steps \\
Seed & 42 \\
\bottomrule
\end{tabular}
\caption{GRPO post-training hyperparameters.}
\label{tab:grpo-hparams}
\end{table}

\paragraph{Supervised Finetuning.}

Starting from Olmo-2-1124-7B basemodel and CPT checkpoint, we run multilingual SFT on \textsc{PolyFact} with the configuration in 
Table~\ref{tab:sft-hparams}.

\begin{table}[h]
\centering
\footnotesize
\setlength{\tabcolsep}{4pt}
\renewcommand{\arraystretch}{1.05}
\begin{tabular}{@{}ll@{}}
\toprule
\textbf{Parameter} & \textbf{Value} \\
\midrule
Dataset & \textsc{PolyFact} \\
Languages per fact ($L$) & 12 \\
Epochs & 1 \\
Facts per device batch & 8 \\
Gradient accumulation & 8 \\
Effective batch (facts) & 64 \\
Learning rate & $2\!\times\!10^{-5}$ \\
LR schedule & cosine, 5\% warmup \\
Weight decay & 0.01 \\
Optimizer & AdamW (fused) \\
Max eval samples/lang & 200 \\
LoRA $r$ / $\alpha$ / dropout & 64 / 128 / 0.05 \\
Precision & bf16 \\
Grad checkpointing & enabled \\
Eval / save frequency & every 200 steps \\
\bottomrule
\end{tabular}
\caption{SFT post-training hyperparameters.}
\label{tab:sft-hparams}
\end{table}

\section{Failure Modes during Cross-Lingual Retrieval}
\label{app:failure-modes}

Performing a layer-rank analysis of the base and GRPO-finetuned model we can get a better understanding of the specific ways in which we have improved performance. The main failure mode of the base model across all languages is producing incoherent answers including randomly guessed numbers, the word ``what" and hybrid combinations of different languages. GRPO is effective in reducing this and contributes most of the gain in performance. Beyond this, the next most common failure modes that we are able to improve upon are largely distinct between languages that do and do not share a Latin script with English. We include Russian in the Latin-script group here: although Cyrillic, it shares enough tokenisation overlap with Latin that it exhibits the same failure mode.  In essence, the base model suffers from some languages' representations being too close to that of English, such that their interlingual pathways are not distinct enough; and some being too separated, such that the routes between them and English are too weak. GRPO fine-tuning helps to reduce both contrasting issues.

\begin{figure}[h]
    \centering
    \includegraphics[width=1\linewidth]{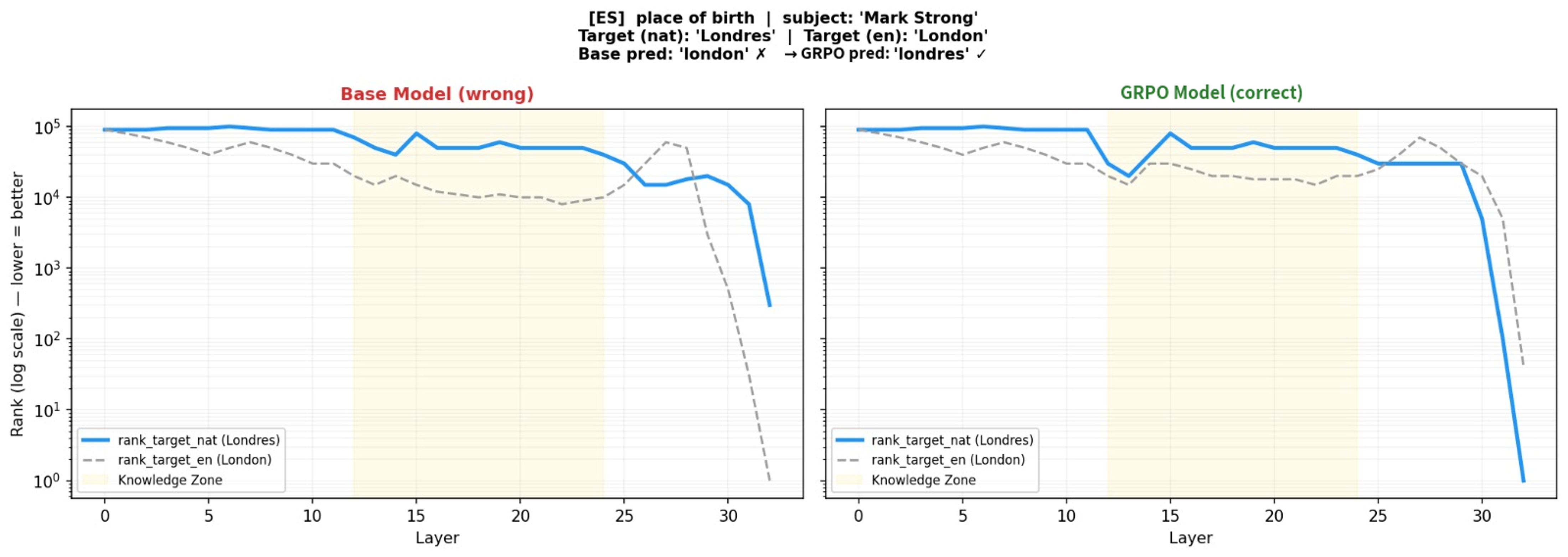}
    \caption{Layer-rank analysis of Base and GRPO finetuned OLMO-2 models for a prompt in latin script language.}
    \label{fig:mark_strong_2}
\end{figure}

Those that do share the same script, and likely similar representations within the model, suffer from responding to prompts with the right answer but in English rather than the target language (Figure \ref{fig:mark_strong_2}). Either the base model cannot find the direct translation for the prompt's answer or its prior training has conditioned it to rely too heavily on English-tied pathways within the Latin-script regions of its representations. The former seems unlikely, particularly given the expected prominence of the answer in the specific example shown. Therefore, our GRPO fine-tuning is able to encourage the model to separate pathways for Latin script languages and respond in the correct language. This unfortunately does not apply to all examples and in fact GRPO produces nearly as many regressions of this type as it does improvements. Though this looks discouraging on the surface, a deeper analysis of where these failures occur reveals much about the base model and our fine-tuning process. 

The KLAR dataset is structured into \textit{relations} that group questions from the same fields. We break these relations down into two groups, those with and without proper-noun answers (jurisdiction, manufacturer, owned-by etc). Given the global prevalence of English-language proper nouns it is understandable that many of these proper-nouns originate in the English language and may be known as such, even if they have native translations. In some instances this will naturally aid the answering of these questions: target-language answers are often similar or identical to the English forms the model can retrieve more easily. Proper-noun relations account for 61\%, 76\%, and 73\% of base-to-GRPO regressions in Russian, French, and Spanish respectively. While the prevalence of English proper-nouns in the base model's pretraining corpus likely helps it score well on the low-hanging lexical cognates, it also pushes many answers into the wrong language failure state — a tendency our finetuning may have inadvertently reinforced. Wikipedia is well known for exhibiting this bias, where entity labels for people, companies, and many places default to English or Latin-script forms even on non-English Wikipedias, because Wikidata's canonical label is often English and localisers haven't filled in every language \citep{kaffee2017glimpse}. Finetuning on \textsc{PolyFact} therefore strengthens this bias, worsening the wrong-language failure mode on proper-noun prompts. 

\begin{figure}
    \centering
    \includegraphics[width=1\linewidth]{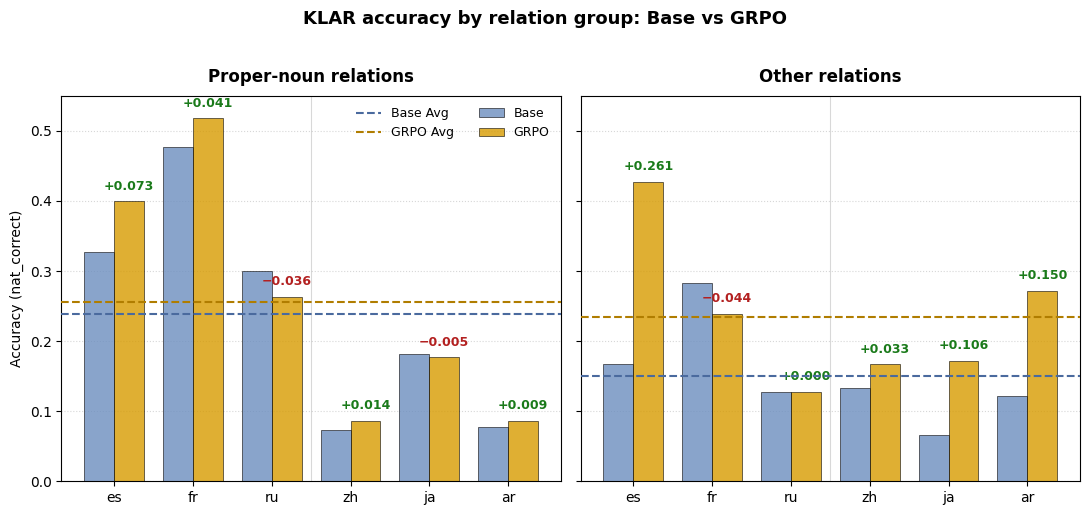}
    \caption{KLAR improvement by relation type. Proper-noun relations account for 55\% of the KLAR dataset.}
    \label{fig:relation_comp}
\end{figure}

This points us to the notion that GRPO fine-tuning is actually stronger than our initial metrics suggest. The gains GRPO actually delivers on language abstraction are partly masked in the aggregate by regressions concentrated in proper-noun prompts where dataset bias works against us. Figure \ref{fig:relation_comp} makes this visible: improvements on common-noun (non-proper-noun) relations are substantially larger across the board, including in non-Latin-script languages. 

We highlight the regression of Russian performance after our fine-tuning, as it is a slight anomaly though we cannot fully explain it. Some gains come from cleaning up incoherent answers, but the overall result is dominated by regressions in the same failure mode as the Latin based languages. This complicates our script-based framing but the aforementioned the English-leak failure mode reaches it too. What we cannot fully explain is its magnitude: Russian is the only language to net-regress on KLAR, and does so more severely than fully Latin-script French, despite sharing the underlying mechanism.

In languages that do not share the same script or have this overlap the failure mode for questions not related to proper-nouns is qualitatively different. These languages perform much worse in general evaluation, in part because of weaker factual retrieval: responses are more often in the correct language but factually wrong (Figure \ref{fig:zh_plot_flat}). The model's representations for these languages are clearly distinct from those of English and its language-agnostic regions. This separation means the base model answers in the correct language but the cross-lingual transfer of factual content from the English/agnostic knowledge region into the language-specific output pathway fails. Our GRPO fine-tuning strengthens the pathways between these distinct languages and more commonly allows the model to answer in the correct language with the correct information.

Our GRPO fine-tuning uses a reward function that requires both the correctness of the answer and its production in the target language to assign credit. This is what enables us to correct both failure modes, pushing the model towards answering correctly in the right language from either direction relative to the target. But there is no notion of partial credit and given the distinct failure modes (correct language, wrong information vs wrong language, correct information) it seems that there may be specific benefits tailored to each language type to including this in future iterations. Similarly, an altered training dataset could yield further gains. The reward rewards correctness, and the canonical correct answer is often English, so the reward effectively trains in the English-leak failure on proper-noun prompts The aforementioned biases of the \textsc{PolyFact} database could be corrected to favour more consistently native proper nouns and we would likely see the true gains of this method unhampered by the discussed regressions. 



\section{\textsc{PolyFact} Dataset.}
\label{app:polyfact}

\subsection{Detailed Construction.}
We describe the full pipeline used to construct \textsc{PolyFact}, a parallel
multilingual multiple-choice QA dataset grounded in Wikidata.

\textit{Source data and relation selection.} We start from the full Wikidata
truthy-triples dump and retain only triples whose property belongs to a
curated set of 22 factual relations, selected to cover stable, unambiguous
facts suitable for MCQ generation. These span geography (\textit{capital},
\textit{country}, \textit{continent}, \textit{official language},
\textit{currency}, \textit{shares border with}), biography (\textit{country of
citizenship}, \textit{place of birth}, \textit{place of death},
\textit{educated at}, \textit{employer}), creative works and media
(\textit{author}, \textit{director}, \textit{creator}, \textit{developer},
\textit{genre}, \textit{country of origin}, \textit{language of work or
name}, \textit{platform}), and organizational or cultural relations
(\textit{manufacturer}, \textit{architect}, \textit{discoverer or inventor}).
Three of these (\textit{capital}, \textit{shares border with},
\textit{platform}) do not survive the label-coverage filter below in
sufficient numbers, so the generated corpus spans 19 relations; quality
verification (Appendix~\ref{app:polyfact-quality}) later removes five more,
and the released corpus spans 14.

\textit{Multilingual label extraction.} For every subject, property, and
object entity, we extract labels in twelve typologically and geographically
diverse languages: English, German, Indonesian, Portuguese, Arabic, Bengali,
Swahili, Spanish, Russian, French, Japanese, and Chinese. Labels are collected
primarily by streaming the compressed Wikidata JSON dump, with missing entries
backfilled through the Wikidata \texttt{wbgetentities} API in batches of 50
IDs. We additionally extract each entity's \texttt{instance of} (P31) types
from the dump for distractor selection.

\textit{Triple filtering and distractor construction.}
We retain only triples whose subject and object have labels in a sufficient
subset of the twelve target languages. For each surviving triple, we sample
three distractors from objects that appear with the same property elsewhere in
the corpus, so distractors are plausible candidates for the relation (e.g.,
distractors for a \emph{place of death} question are drawn from other
Wikidata-recorded places of death). To prevent surface-level shortcuts,
distractors must additionally share at least one entity type with the gold
object --- via Wikidata's \texttt{instance of} property (P31), e.g.\ both
being cities --- and match its English label length within a small tolerance.
We discard any fact whose four options are not all distinct under case- and
whitespace-normalised comparison. (These initially sampled distractors are
later replaced wholesale by the resampling stage of the quality pipeline,
Appendix~\ref{app:polyfact-quality}, which supersedes the type- and
length-matching constraints.)

\textit{Balanced sampling.} To prevent high-frequency relations from
dominating the corpus, we apply round-robin sampling across properties: on
each pass, one fact is drawn from each property's shuffled pool, cycling until
100{,}113 candidate facts have been selected. This produces a near-uniform
distribution over relations regardless of the underlying frequency skew in
Wikidata, up to pool exhaustion for the rarest properties (\textit{author},
\textit{currency}).

\textit{Multilingual MCQ generation.} For each selected fact, we generate a
parallel MCQ bundle covering all twelve languages using gemma-3-27b-it \cite{gemmateam2025gemma3} served
via vLLM. Entity and property labels missing from Wikidata are prefilled using
a dedicated low-temperature translation prompt (temperature $0$, max 32
tokens) conditioned on the relation context, with results cached across the
batch to amortize cost. Given the localized subject, relation, gold answer,
and three distractors, the model generates a question in the target language
under explicit instructions to reuse the provided option strings verbatim
(temperature $0.1$, top-$p$ $0.9$, max 192 tokens). The predicted
\texttt{answer\_text} is resolved against the option set via exact match and,
on failure, case- and whitespace-normalized string matching against both the
options and the known gold label.

\textit{Validity and parallelism.} Each generated MCQ is validated for
well-formedness: exactly four distinct non-empty options, a non-empty question
string, and an \texttt{answer\_text} that resolves to one of the options. A
fact is retained only if generation succeeds for \emph{all} twelve languages;
partial bundles are discarded entirely. This strict all-or-nothing criterion
guarantees that every fact in \textsc{PolyFact} is fully parallel, so
cross-lingual performance differences can be attributed to model behavior
rather than to variation in underlying content, question difficulty, or
answer set composition.

\textit{Splits.} Splits are applied at the fact level, so all twelve language
versions of a fact remain in the same partition. After the quality pipeline
below, the released corpus contains 56{,}324 training, 444 validation, and
2{,}039 test facts. The test split is deliberately the larger of the two
evaluation splits to ensure
statistical significance for the paper's main comparisons.

\subsection{Quality Verification and Cleaning.}
\label{app:polyfact-quality}
The candidate corpus passes through a four-stage quality pipeline; in total,
40{,}822 of the 100{,}113 candidate facts (40.8\%) are removed and every
surviving evaluation item is individually verified.

\textit{(1) Relation-level audit.} We labelled 300 facts per language across
all 12 languages (3{,}600 items) with an LLM-as-judge (GPT-4o with web-search
grounding) and independently human-labelled 100 items per language across
seven languages (700 items), using a three-label rubric (correct~/ ambiguous~/
incorrect). Inter-judge agreement is 91.0\% overall, ranging from 86\%
(Arabic) to 96\% (English) per language (Table~\ref{tab:polyfact_iaa}). The
judge finds an overall ambiguity rate of 13.3\%, concentrated in three
relations --- \textit{country of origin} (35\%), \textit{place of birth}
(25\%), and \textit{genre} (25\%) --- driven by shared person names and the
inherent subjectivity of genre (Table~\ref{tab:wikifact_quality}). These three
relations are removed. We additionally remove \textit{employer}, whose
many-jobs-per-person structure makes a single gold arbitrary and which
contained every known split-integrity defect of the candidate corpus, and
\textit{currency}, none of whose answer entities carry Wikidata labels in all
twelve languages, leaving its labels unverifiable.

\textit{(2) Label audit.} Every answer label is checked against the union of
Wikidata's labels, aliases, language variants, and Wikipedia sitelinks. Labels
this audit cannot attest are risk-stratified by romanization: phonetic
renderings of the English name are low-risk, while semantic translations can
denote the wrong referent. We use Claude Opus-5 to manually review the entire high-risk stratum
--- 7{,}708 labels --- finding 275 wrong (3.6\%; worst in Swahili at 6.9\%,
typically a proper name that is also a common noun translated by meaning). The
207 affected entities are removed globally, since distractors draw from the
same answer pool. Facts with transliteration-corrupted labels (a Latin
fragment fused inside native script, e.g.\ Bengali) are removed rather than
rewritten, as Wikidata offers no ground truth to repair from.

\textit{(3) Distractor resampling.} An audit of the
generated corpus found that its distractors leak an entity-frequency prior: a
model-free baseline that picks the option occurring most often as a gold
answer elsewhere scores 69.97\% on the test split (chance 25\%), and because
this prior is entity-based it is language-independent, so exploiting it would
inflate exactly the cross-lingual consistency metrics we study. We therefore
resample \emph{all} distractors from the entities that serve as gold answers
for the same relation, in proportion to their gold frequency, making the four
options exchangeable draws from a single distribution; the baseline falls to
29.15\%.

\textit{(4) Exhaustive evaluation-split census.} We used Claude Opus-5 to individually review all 27{,}753 non-English questions of the
(pre-drop) 2{,}523-fact test split, which were
judged against the English reference for subject and property fidelity, and
every English question was fact-checked against its stored gold.
Translation defect rates are 3.85\% (95\% CI 3.63--4.08) on test and 3.54\%
(3.06--4.10) on validation, dominated by two systematic generation bugs:
obscure place or work names rendered as if they were \emph{languages}, and
proper names translated by their common-noun meaning. The English question
itself was flagged as contradicting the stored gold in 4.08\% (test) and
2.03\% (validation) of cases, typically via an invented year or medium
(``the 1979 film Darr''; the film is from 1993); editorial adjudication
confirmed all but one flag, overturning one English and one translation flag
as false positives. Confirmed defects were \emph{repaired}, not dropped: target-language
questions were re-translated from English by per-language models given real
same-relation corpus questions as style anchors, then accepted only after
mechanical gates for answer leakage, distractor leakage, script, and length;
English conflicts were editorially corrected by removing the invented detail.
One template-level bug (784 Arabic \textit{country of citizenship} questions
rendering ``country'' as ``municipality''), found during this review, was
repaired corpus-wide --- across all three splits --- by substitution to the
corpus's own majority template.

\textit{(5) Independent item-quality audit and final drop.} The census above
measures fidelity to the English reference and can therefore miss defects
English fidelity cannot reveal. After repairs, a second, independent LLM
judge (Claude Opus-5) re-audited every post-repair test item for question, label, and
validity issues against world knowledge (giving confidence labels). Wrong-entity flags whose disputed
name matches an official Wikidata label or alias for that language were
rescued as false positives (117 of 302) and a stratified human recheck of
28 flags confirmed 26--27, adjudicating the two contested cases via
Wikidata. The 484 test facts still carrying a high-confidence flag were then
\emph{removed} -- 309 of them defects the fidelity census was structurally
blind to -- leaving 2{,}039 test facts that passed both audits. In addition the includes \texttt{n\_langs\_verified} label-attestation count,
so users can apply stricter subsets than we do. All reviews were
LLM-assisted rather than native-speaker-verified, with human spot-checks at
each stage.

\begin{table}[t]
\centering
\small
\begin{tabular}{lrr}
\toprule
Language & $N$ & Agreement \\
\midrule
English (en)              & 100 & 96.0\% \\
French (fr)               & 100 & 93.0\% \\
German (de)               & 100 & 92.4\% \\
Spanish (es)              & 100 & 92.0\% \\
Russian (ru)  & 100 & 87.0\% \\
Chinese (zh)  & 100 & 91.9\% \\
Arabic (ar)               & 100 & 86.0\% \\
\midrule
Overall                   & 700 & 91.0\% \\
\bottomrule
\end{tabular}
\caption{LLM-human inter-judge agreement on the relation-level audit sample
(100 items per language). $N$ is the number of items with both an LLM-judge
label and an independent human label.}
\label{tab:polyfact_iaa}
\end{table}

\begin{table}[t]
\centering
\small
\begin{tabular}{lrr}
\toprule
Relation & $N$ & Ambig.\ \% \\
\midrule
\multicolumn{3}{l}{\emph{Removed from \textsc{PolyFact}}} \\
country of origin       & 168 & 35 \\
genre                   & 240 & 25 \\
place of birth          & 216 & 25 \\
currency$^\dagger$      &  36 & 14 \\
employer$^\dagger$      & 216 & 11 \\
\midrule
\multicolumn{3}{l}{\emph{Retained}} \\
director                & 192 & 20 \\
continent               & 216 & 20 \\
language of work or name& 252 & 18 \\
educated at             & 240 & 17 \\
architect               & 180 &  9 \\
official language       & 168 &  8 \\
creator                 & 192 &  8 \\
place of death          & 216 &  8 \\
developer               & 192 &  7 \\
country                 & 192 &  6 \\
country of citizenship  & 216 &  5 \\
author                  &  60 &  3 \\
discoverer or inventor  & 216 &  3 \\
manufacturer            & 192 &  2 \\
\bottomrule
\end{tabular}
\caption{Per-relation ambiguity rates from the 3{,}600-item LLM-judge audit
(300 items per language across 12 languages), covering all 19 relations of
the generated corpus. The top three relations are removed for ambiguity;
$^\dagger$\textit{currency} and \textit{employer} are removed on separate
grounds (unverifiable labels; many-to-one gold structure and split-integrity
defects --- see text). The released corpus spans the 14 retained relations.}
\label{tab:wikifact_quality}
\end{table}

\section{Additional Results}

\subsection{Mechanistic Analysis via LAHIS}

\begin{figure}[th!]
    \centering
    \includegraphics[width=1\linewidth]{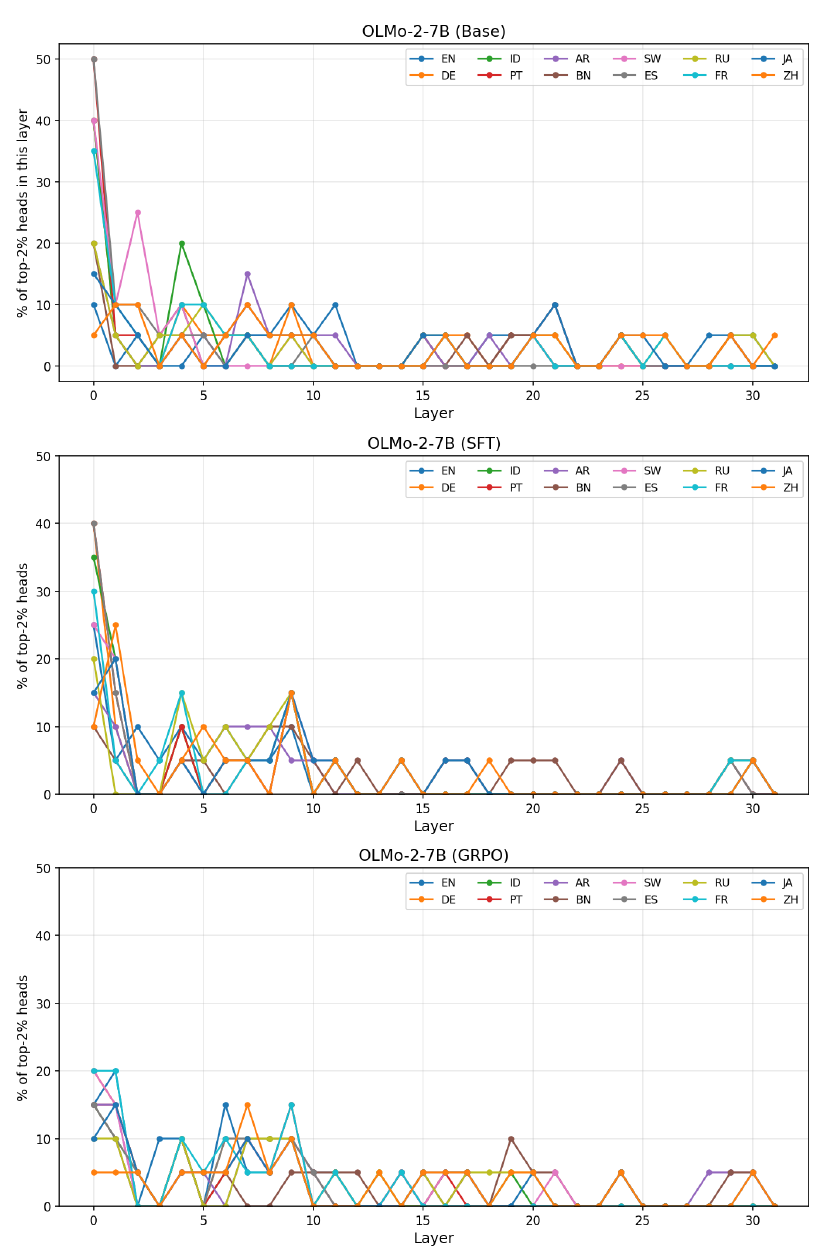}
    \caption{Percentage of language-important attention heads in OLMo-2-1124-7B across Base, SFT and GRPO.}
    \label{fig:head_distribution_olmo}
\end{figure}

\paragraph{Cross-lingual head sharing.}
Head overlap between language pairs increases substantially after both finetuning methods (as visualized in Figure~\ref{fig:head_overlap}). SFT produces the strongest gains in pairwise overlap, particularly among Indo-European languages (e.g.\ DE--FR: 25\% $\rightarrow$ 90\%, DE--ID: 35\% $\rightarrow$ 85\%). GRPO also increases overlap but more moderately, with notable gains for typologically distant pairs such as JA--ZH (50\% $\rightarrow$ 80\%). This indicates that SFT can  encourage shared routing in attention layers across languages, while GRPO's key effect lies in redistributing processing deeper into the network (as seen in the LAPE results) rather than maximising head sharing.

\paragraph{Language processing shifts deeper into the network.}
In the base model up to 50\% of language-important heads are concentrated in layer 0, suggesting that language-specific processing is established early in the network. Both finetuning methods redistribute these heads: under SFT, the peak drops to around 40\% with a modest spread across early layers, while GRPO reduces it further to approximately 20\% and distributes language-important heads more broadly across layers 0--10 (Figure~\ref{fig:head_distribution_olmo}). This pattern suggests that GRPO is associated with a more substantial reorganisation of early language processing than SFT. Together with the overlap analysis, the results indicate qualitatively different effects of the two methods: SFT primarily increases cross-lingual head sharing, whereas GRPO more strongly changes the layer-wise organisation of language-important heads, consistent with later and less concentrated language-specific processing. As LAHIS is correlational, however, these patterns do not establish that the observed redistribution causes the behavioural improvements. Additional results are provided in Appendix~\ref{app:lahis-extended}.

\subsection{LAHIS}
\label{app:lahis-extended}

Despite similar overlap patterns, the two methods differ in how much they alter the base 
model's routing. Under SFT, between 45\% and 75\% of language-important heads change 
relative to the base, while GRPO replaces between 70\% and 85\% 
(Figure~\ref{fig:head_stability}). Per-language delta maps for both models are provided in Appendix~\ref{app:delta}. Pairwise head overlap for the three Qwen-2.5-7B variants is visualized in Figure \ref{fig:qwen-pairwise-head-overlap-qwen}.

\begin{figure}[t]
    \centering
    \includegraphics[width=\columnwidth]{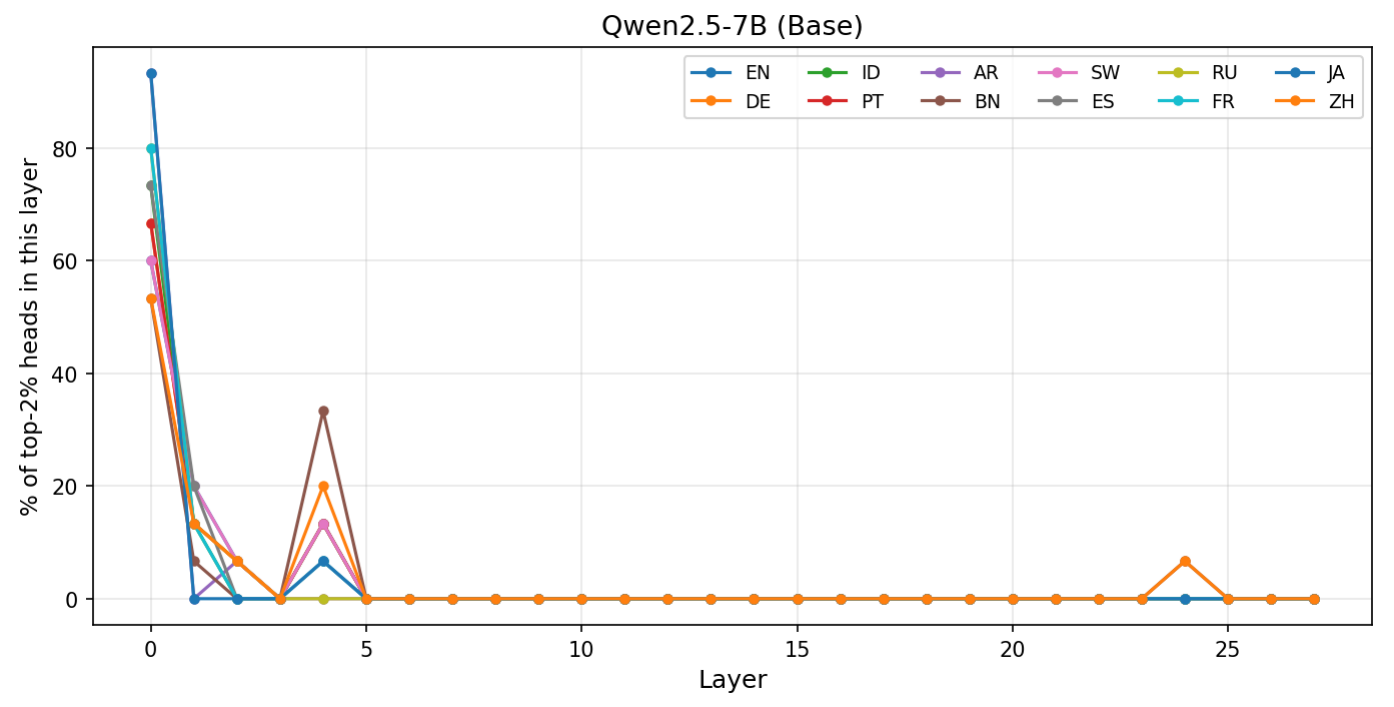}
    \vspace{-0.6em}

    \includegraphics[width=\columnwidth]{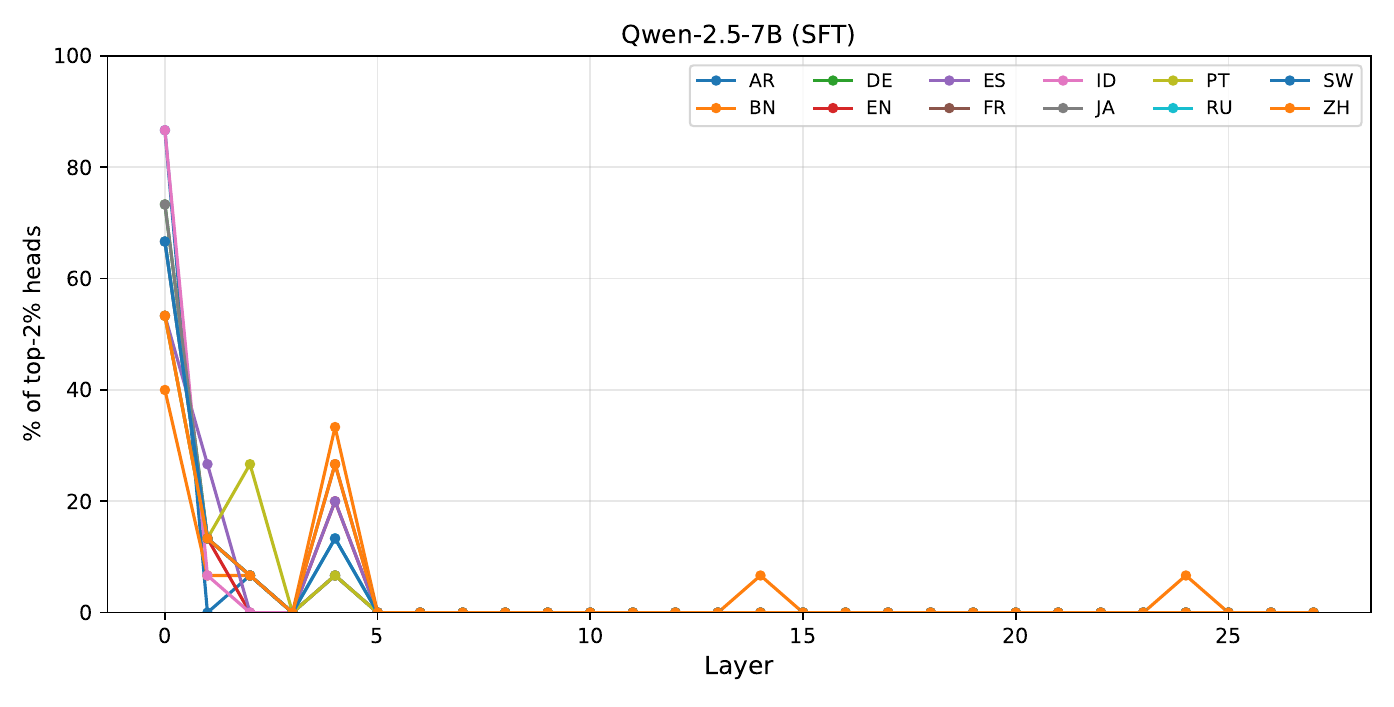}
    \vspace{-0.6em}

    \includegraphics[width=\columnwidth]{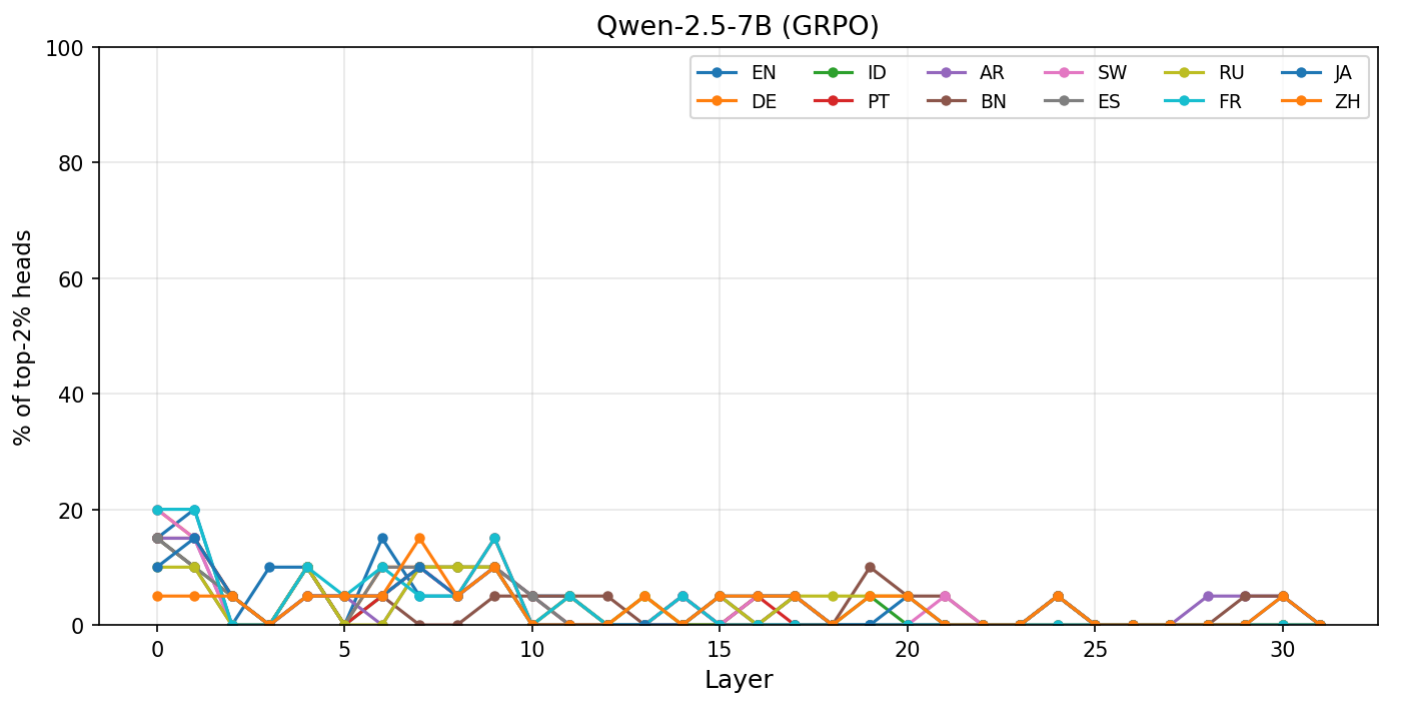}

    \caption{Percentage of language-important attention heads per layer for Qwen-2.5-7B before and after post-training. GRPO reduces the concentration of language-specific heads in the earliest layers and distributes language routing more broadly across the network.}
    \label{fig:qwen-head-distribution}
\end{figure}

\begin{figure}[h]
\centering
\begin{subfigure}[t]{0.31\linewidth}
    \centering
    \includegraphics[height=2.6cm, trim=100pt 0 200pt 0, clip]{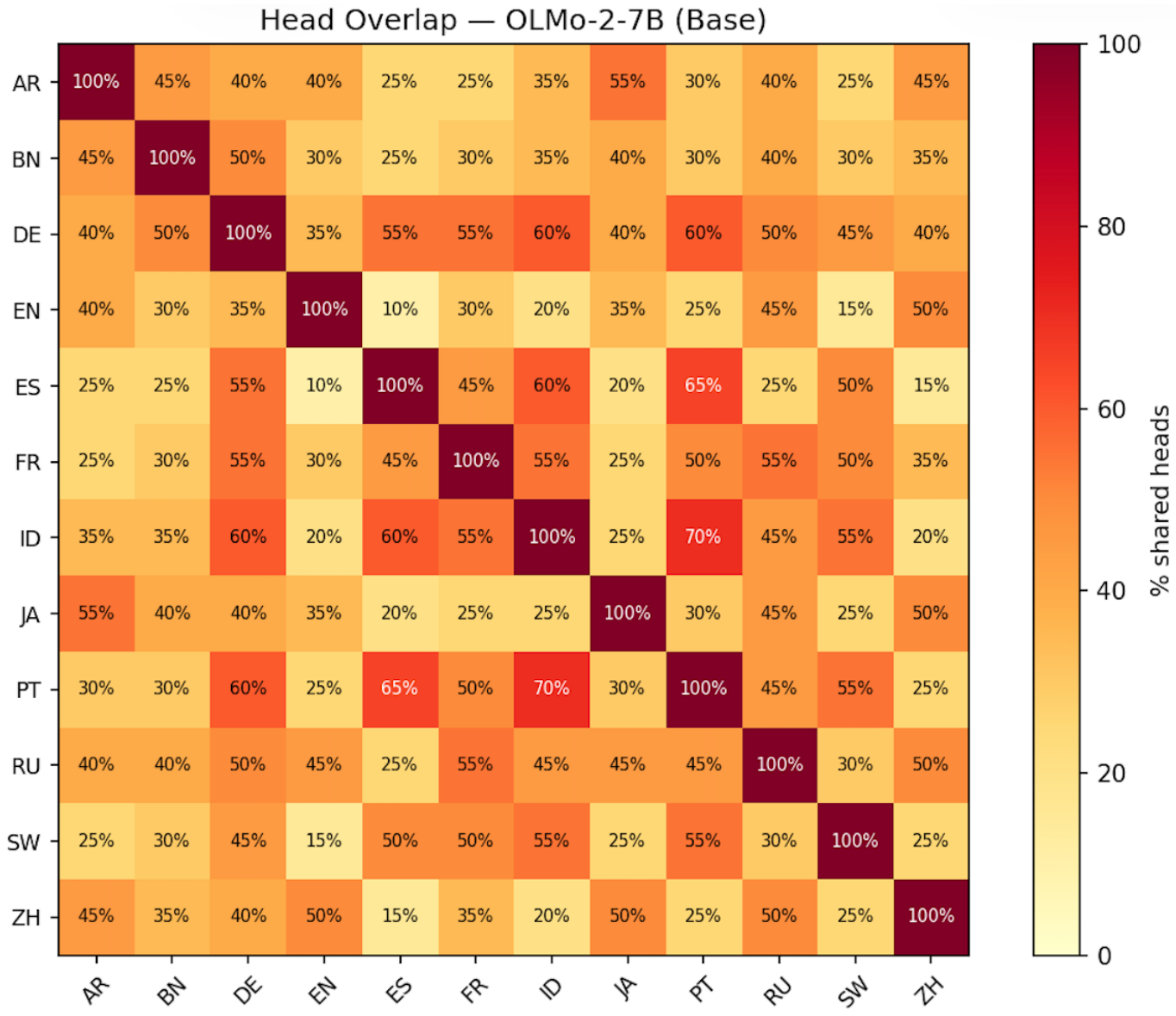}
    \caption{Base}
\end{subfigure}
\hfill
\begin{subfigure}[t]{0.31\linewidth}
    \centering
    \includegraphics[height=2.6cm, trim=100pt 0 200pt 0, clip]{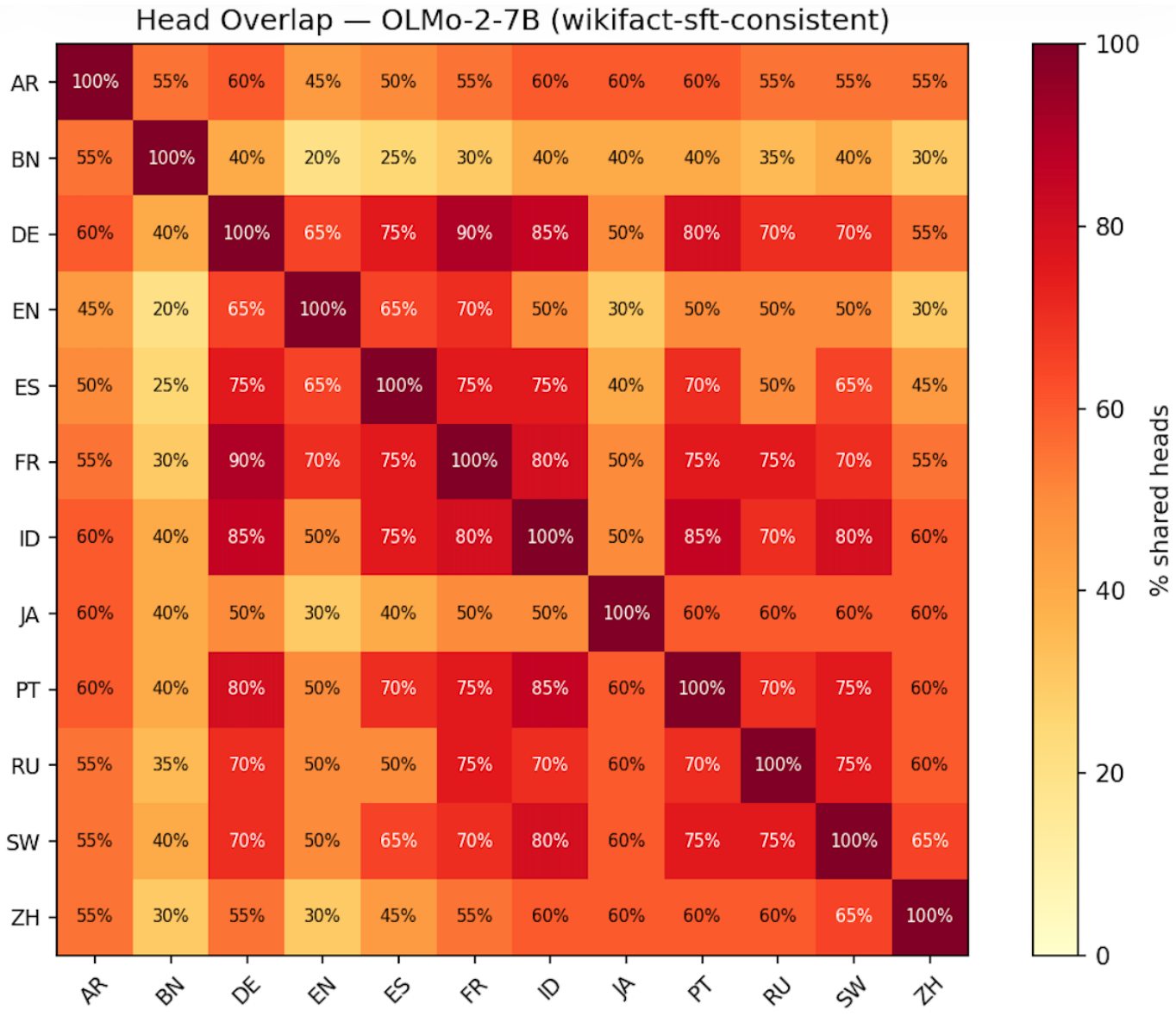}
    \caption{SFT}
\end{subfigure}
\hfill
\begin{subfigure}[t]{0.35\linewidth}
    \centering
    \includegraphics[height=2.6cm, trim=100pt 0 90pt 0, clip]{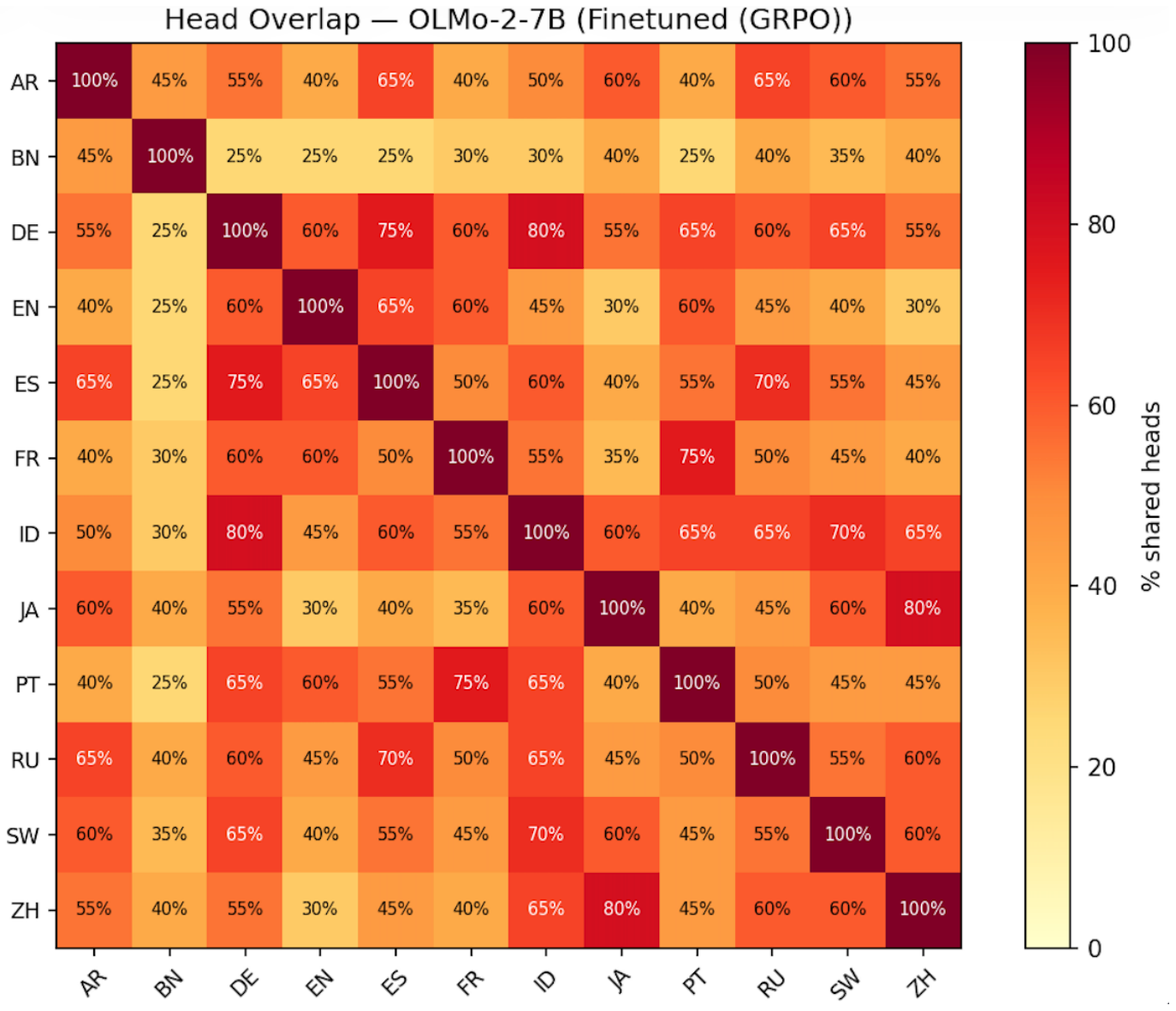}
    \caption{GRPO}
\end{subfigure}
\caption{Pairwise head overlap across language pairs for Base, SFT-, and GRPO-finetuned models. 
}
\label{fig:head_overlap}
\end{figure}

\begin{figure}[t]
\centering
\begin{subfigure}[t]{0.48\columnwidth}
\includegraphics[width=\linewidth]{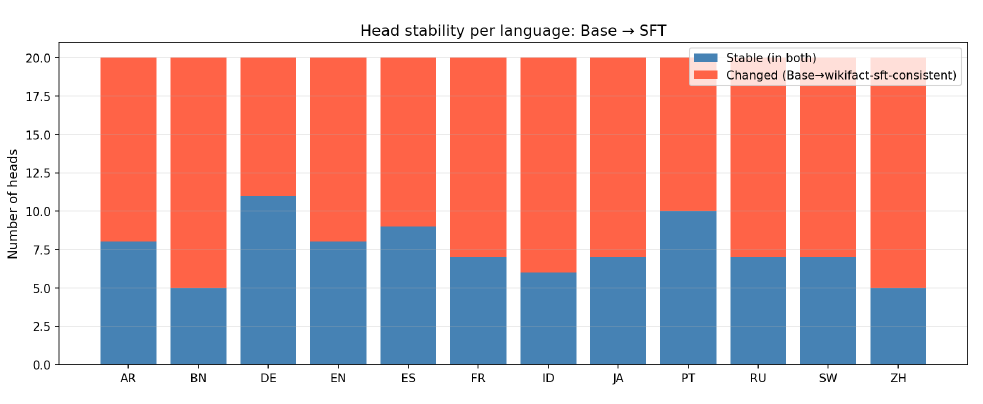}
\caption{Base $\rightarrow$ SFT}
\end{subfigure}
\hfill
\begin{subfigure}[t]{0.48\columnwidth}
\includegraphics[width=\linewidth]{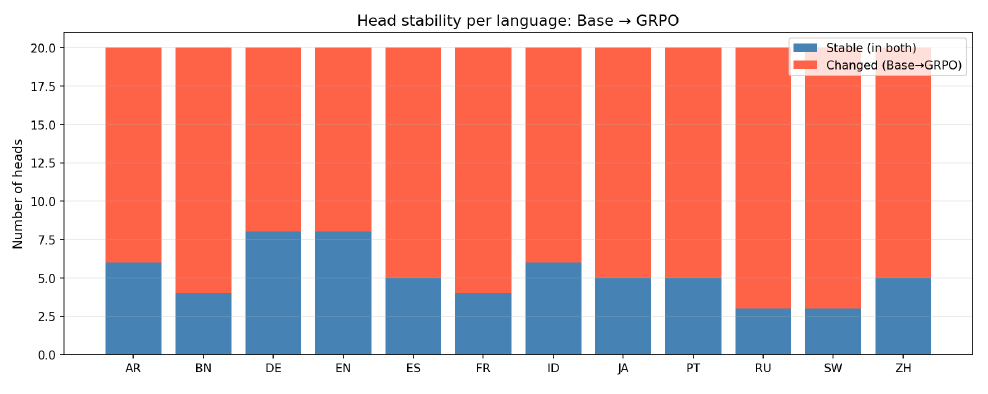}
\caption{Base $\rightarrow$ GRPO}
\end{subfigure}
\caption{Number of stable (blue) and changed (orange) language-important heads per 
language. 
}
\label{fig:head_stability}
\end{figure}

\begin{figure*}[t]
    \centering
    \begin{subfigure}[t]{0.32\textwidth}
        \centering
        \includegraphics[width=\linewidth]{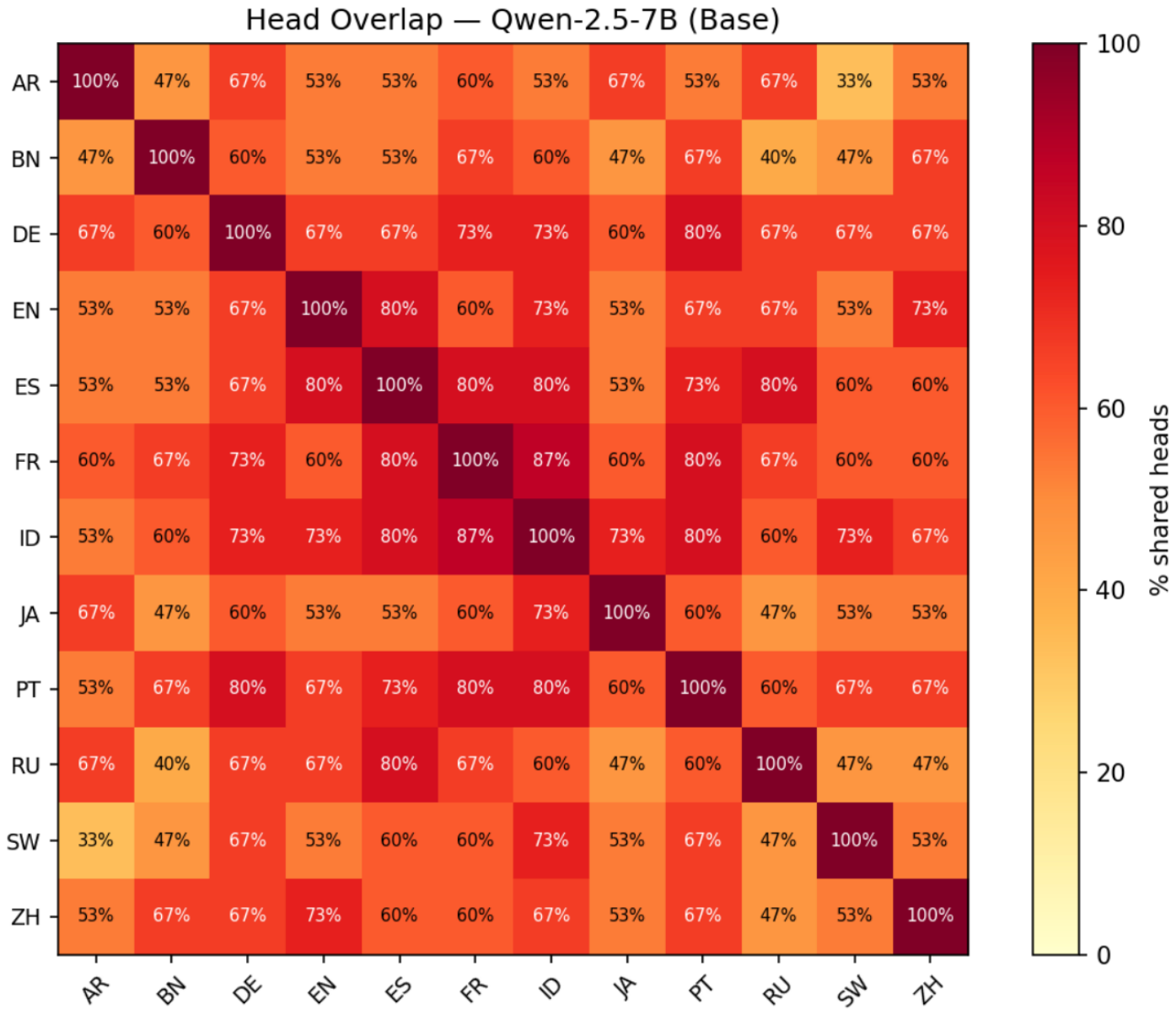}
        \caption{Base}
    \end{subfigure}
    \hfill
    \begin{subfigure}[t]{0.27\textwidth}
        \centering
        \includegraphics[width=\linewidth]{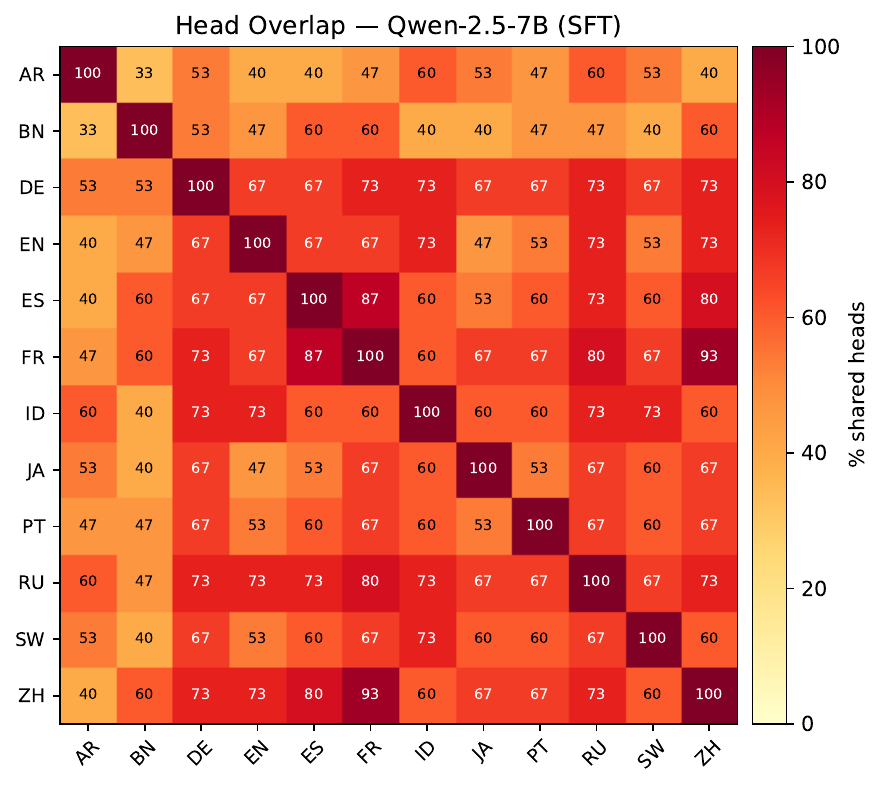}
        \caption{SFT}
    \end{subfigure}
    \hfill
    \begin{subfigure}[t]{0.32\textwidth}
        \centering
        \includegraphics[width=\linewidth]{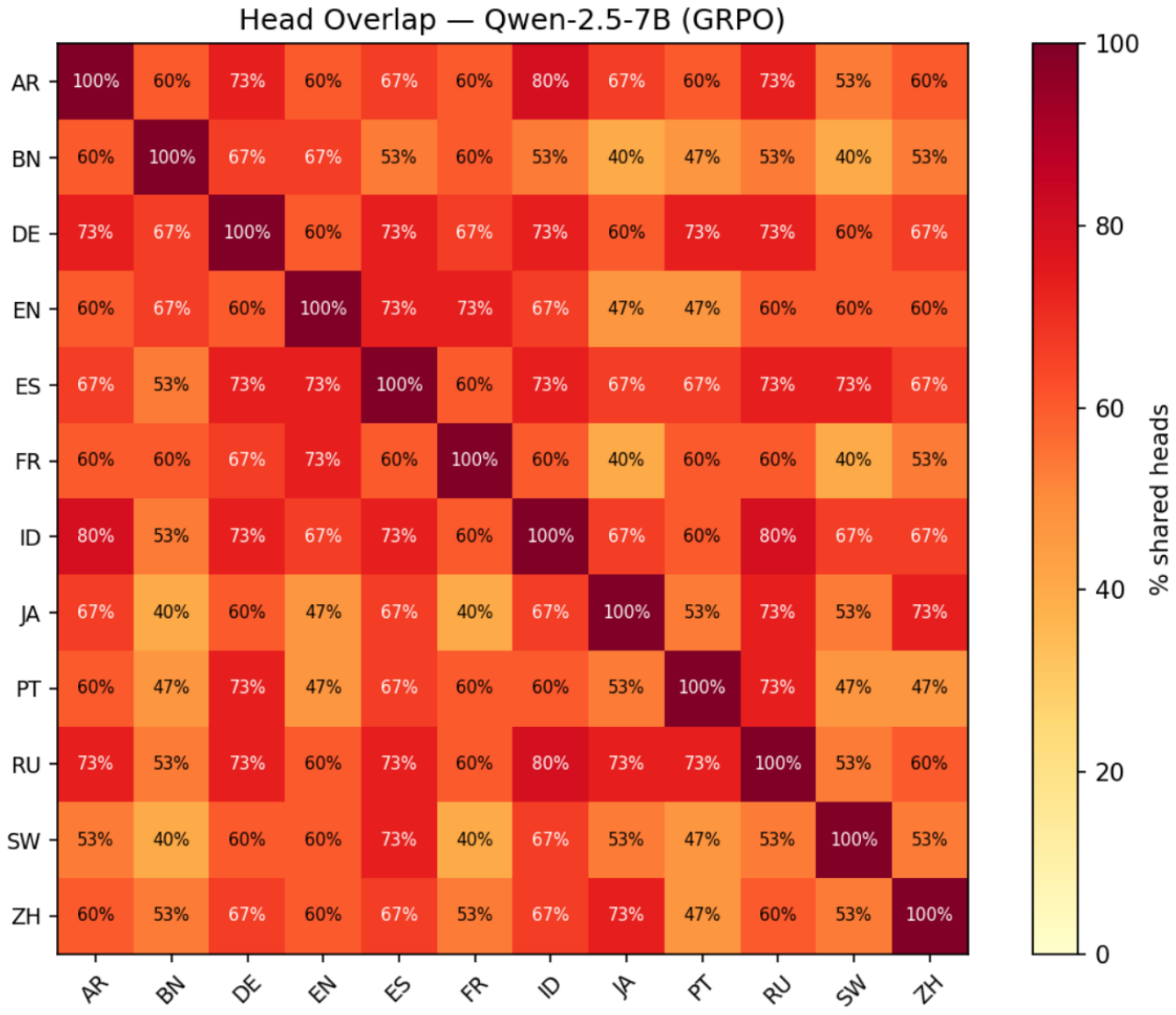}
        \caption{GRPO}
    \end{subfigure}

    \caption{Pairwise overlap of language-important attention heads across language pairs for Qwen-2.5-7B before and after post-training.}
    \label{fig:qwen-pairwise-head-overlap-qwen}
\end{figure*}

\paragraph{Per-language Delta Maps}
\label{app:delta}

Figures~\ref{fig:delta_grpo} and~\ref{fig:delta_sft} show per-language delta maps of 
LAHIS importance scores (finetuned $-$ base) for GRPO and SFT-consistent respectively. 
Both methods predominantly suppress existing language-specific heads rather than 
creating new ones, with the strongest suppression at layer~0.

\subsection{Per-language Neuron Distributions}
\label{app:lape_full}

\begin{figure*}[h]
    \centering
    \includegraphics[width=0.95\textwidth]{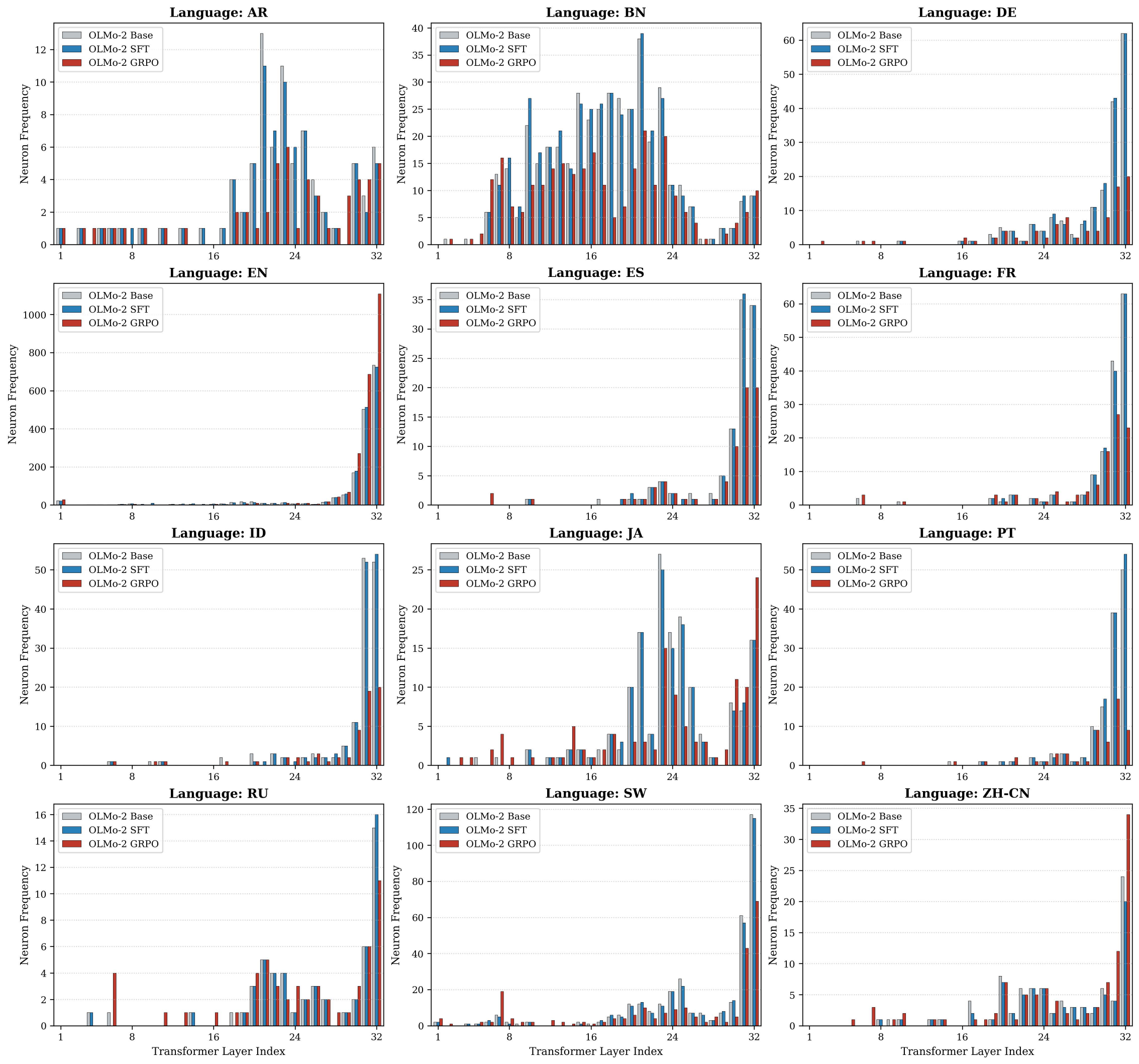}
    \caption{Layer-wise frequency of language-specific neurons for all target languages. Note the distinct English (EN) surge and the redistribution of neurons in non-Latin scripts (AR, JA, ZH-CN) under the GRPO configuration (red).}
    \label{fig:lape_full_grid}
\end{figure*}

\begin{figure*}[h]
    \centering
    \includegraphics[width=\linewidth]{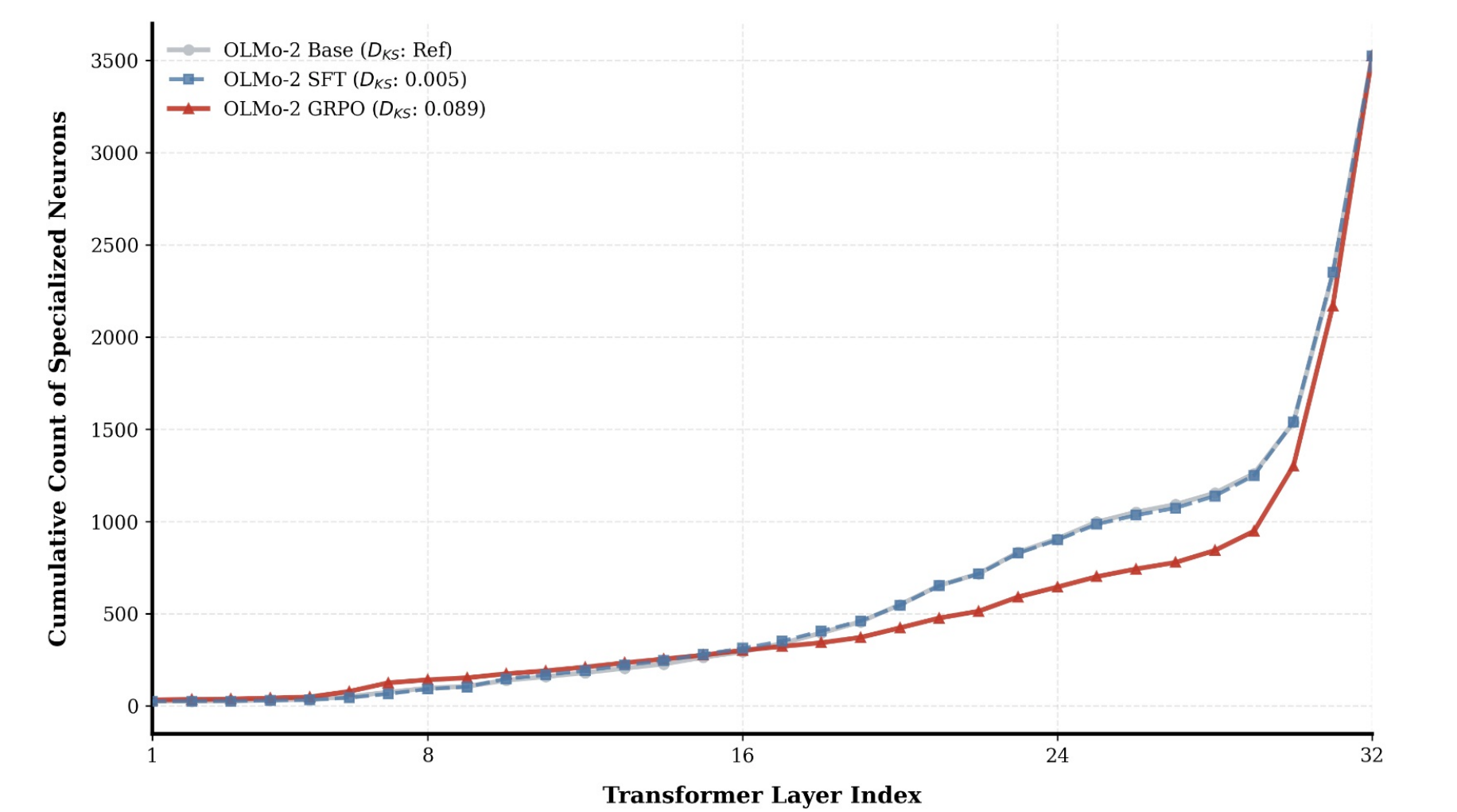}
    \caption{Empirical Cumulative Distribution Function (ECDF) of specialised neuron discovery across layers. The significant $D_{KS}$ shift in GRPO indicates a deferral of linguistic specialisation to later layers.}
    \label{fig:lape_ecdf}
\end{figure*}

\begin{figure*}[b]
\centering
\includegraphics[width=\textwidth]{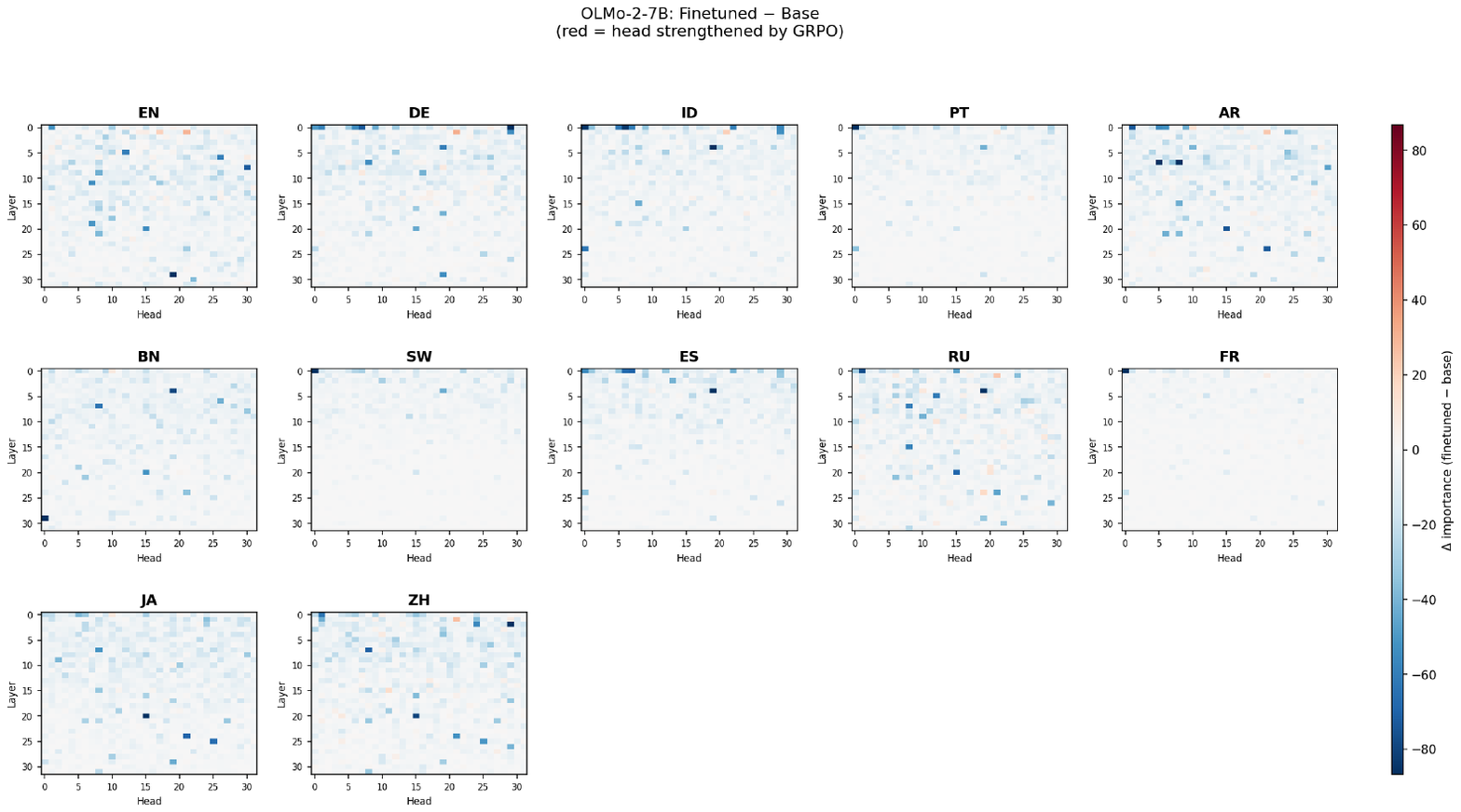}
\caption{Delta heatmap (GRPO $-$ base). Blue indicates weakened heads, red indicates 
strengthened heads.}
\label{fig:delta_grpo}
\end{figure*}

\begin{figure*}[b]
\centering
\includegraphics[width=\textwidth]{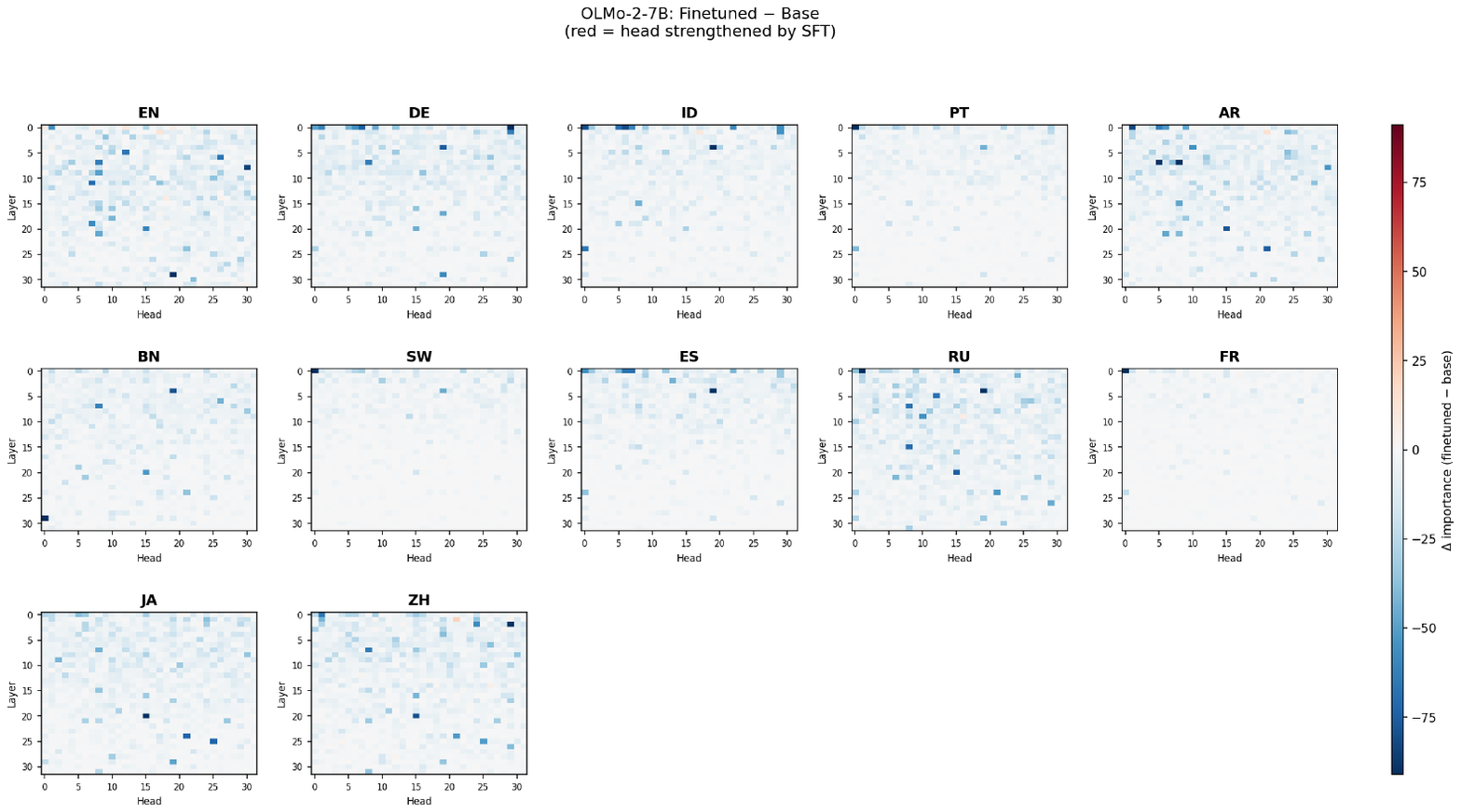}
\caption{Delta heatmap (SFT-consistent $-$ base). Pattern closely mirrors GRPO, 
suggesting both methods suppress similar heads.}
\label{fig:delta_sft}
\end{figure*}

Figure~\ref{fig:lape_full_grid} provides a detailed visualisation of the language activation probability entropy (LAPE) results, showing the frequency of specialised neurons per layer for all twelve target languages across the Base, SFT, and GRPO configurations. Figure~\ref{fig:lape_ecdf} provides the empirical cumulative distribution functions for the base, SFT and GRPO configurations, summed over all languages.

\subsection{Per-language Performance}
\label{app:per-lang-perf}

\paragraph{\textsc{PolyFact}:} See Table \ref{tab:wikifact_appendix}.

\paragraph{KLAR:}See Table \ref{tab:klar_appendix}.

\paragraph{Global-MMLU.}
See Table \ref{tab:global_mmlu_appendix}.


\begin{table*}[t]
\centering
\scriptsize
\setlength{\tabcolsep}{3pt}
\begin{tabular}{lcccccccccccc}
\toprule
Model 
& en & de & es & fr & pt & id & ru & zh & ar & ja & sw & bn \\
\midrule
\multicolumn{13}{l}{\textit{OLMo-2-1124-7B}} \\
\midrule
Baseline           & 75.63 & 56.25 & 56.89 & 56.94 & 56.65 & 52.13 & 36.73 & 28.00 & 25.65 & 27.95 & 34.13 & 26.14 \\
Aligned            & 74.01 & 56.94 & 55.52 & 58.26 & 55.81 & 50.32 & 37.76 & 27.71 & 25.36 & 27.66 & 37.67 & 25.45 \\
\cmidrule(lr){1-13}
GRPO               & 76.31 & 61.21 & 61.65 & 62.14 & 59.20 & 55.96 & 38.40 & 30.01 & 26.58 & 29.77 & 34.43 & 26.09 \\
Aligned + GRPO     & 75.92 & 61.16 & 61.16 & 63.07 & 59.34 & 55.27 & 41.39 & 30.26 & 28.54 & 30.06 & 39.19 & 26.73 \\
\cmidrule(lr){1-13}
SFT                & 77.88 & 60.72 & 65.67 & 63.85 & 62.48 & 58.85 & 39.23 & 30.41 & 27.32 & 29.87 & 37.86 & 26.68 \\
Aligned + SFT      & 77.98 & 63.85 & 67.39 & 66.01 & 62.82 & 60.03 & 42.82 & 29.82 & 28.35 & 31.00 & 42.82 & 27.32 \\
\cmidrule(lr){1-13}
DCO                & 71.85 & 55.66 & 59.74 & 56.89 & 55.57 & 51.69 & 41.69 & 30.36 & 26.63 & 30.21 & 39.23 & 27.46 \\
Aligned + DCO      & 70.72 & 57.48 & 59.54 & 59.10 & 56.01 & 52.62 & 41.49 & 30.01 & 27.32 & 29.18 & 41.83 & 25.94 \\
\cmidrule(lr){1-13}
CM-Align           & 76.70 & 62.04 & 62.82 & 62.82 & 59.74 & 54.73 & 39.77 & 28.15 & 26.43 & 30.16 & 35.31 & 25.75 \\
\midrule\midrule
\multicolumn{13}{l}{\textit{Qwen-2.5-7B}} \\
\midrule
Baseline           & 68.61 & 60.62 & 57.82 & 60.91 & 58.46 & 59.15 & 48.50 & 48.60 & 39.58 & 44.92 & 34.72 & 33.06 \\
Aligned            & 70.38 & 52.97 & 51.45 & 51.40 & 54.00 & 50.71 & 41.20 & 42.82 & 34.58 & 42.42 & 35.51 & 32.66 \\
\cmidrule(lr){1-13}
GRPO               & 73.37 & 66.60 & 66.41 & 65.52 & 65.62 & 64.00 & 53.65 & 51.99 & 43.26 & 48.06 & 37.91 & 35.21 \\
Aligned + GRPO     & 73.86 & 58.75 & 58.85 & 56.15 & 57.48 & 61.06 & 43.21 & 47.67 & 32.32 & 41.93 & 40.26 & 32.91 \\
\cmidrule(lr){1-13}
SFT                & 75.43 & 69.74 & 68.27 & 66.60 & 67.93 & 66.01 & 53.21 & 53.41 & 46.00 & 49.53 & 40.46 & 37.91 \\
Aligned + SFT      & 75.09 & 66.41 & 64.49 & 61.65 & 65.08 & 62.78 & 49.78 & 43.45 & 38.79 & 46.54 & 48.60 & 40.51 \\
\cmidrule(lr){1-13}
DCO                & 72.00 & 66.01 & 66.85 & 66.60 & 64.30 & 64.54 & 53.85 & 52.43 & 43.45 & 49.58 & 41.20 & 38.50 \\
Aligned + DCO      & 71.11 & 60.81 & 63.71 & 60.96 & 62.09 & 60.03 & 51.30 & 50.32 & 41.25 & 48.01 & 43.65 & 39.19 \\
\cmidrule(lr){1-13}
CM-Align           & 73.12 & 64.30 & 61.60 & 63.76 & 63.81 & 61.75 & 49.83 & 51.05 & 40.90 & 46.05 & 36.05 & 34.53 \\
\bottomrule
\end{tabular}
\caption{\textsc{PolyFact} accuracy (\%) across languages.}
\label{tab:wikifact_appendix}
\end{table*}

\begin{table*}[t]
\centering
\scriptsize
\setlength{\tabcolsep}{3pt}
\begin{tabular}{lccccccc|cccccccccc}
\toprule
& \multicolumn{7}{c}{Seen} & \multicolumn{10}{c}{Unseen} \\
\cmidrule(lr){2-8} \cmidrule(lr){9-18}
Model 
& en & es & fr & ru & zh & ja & ar 
& ca & el & fa & he & hu & ko & nl & tr & uk & vi \\
\midrule
\multicolumn{18}{l}{\textit{OLMo-2-1124-7B}} \\
\midrule
Baseline           & 62.9 & 22.5 & 40.4 & 20.2 & 9.8 & 8.1 & 7.9 & 14.1 & 19.5 & 2.9 & 14.3 & 9.8 & 9.1 & 24.4 & 14.9 & 10.3 & 13.6 \\
Aligned            & 50.8 & 13.0 & 21.6 & 13.8 & 9.7 & 3.0 & 7.2 & 9.6 & 10.4 & 2.4 & 6.5 & 8.3 & 6.1 & 11.4 & 14.5 & 3.7 & 10.3 \\
\cmidrule(lr){1-18}
GRPO               & 73.8 & 38.8 & 43.6 & 21.8 & 15.2 & 16.8 & 20.7 & 26.7 & 17.4 & 4.4 & 13.9 & 14.2 & 16.6 & 35.9 & 20.1 & 18.6 & 24.3 \\
Aligned + GRPO     & 70.9 & 40.2 & 35.8 & 18.7 & 14.4 & 14.0 & 15.2 & 28.0 & 16.0 & 4.1 & 10.6 & 20.2 & 4.5 & 28.8 & 22.3 & 14.9 & 26.7 \\
\cmidrule(lr){1-18}
SFT                & 67.4 & 23.3 & 39.1 & 23.3 & 9.7 & 8.6 & 9.5 & 14.3 & 16.7 & 3.1 & 14.1 & 10.5 & 5.5 & 27.6 & 16.2 & 9.9 & 11.6 \\
Aligned + SFT      & 52.2 & 11.3 & 15.2 & 10.9 & 9.6 & 3.4 & 6.4 & 8.4 & 12.8 & 2.5 & 3.9 & 8.3 & 3.1 & 8.4 & 16.1 & 4.5 & 9.4 \\
\cmidrule(lr){1-18}
DCO                & 71.3 & 33.2 & 39.1 & 22.9 & 11.2 & 18.3 & 19.9 & 23.9 & 20.1 & 5.4 & 16.5 & 14.7 & 15.5 & 31.4 & 12.6 & 16.5 & 16.5 \\
Aligned + DCO      & 59.1 & 14.7 & 14.7 & 4.7 & 11.5 & 6.1 & 11.5 & 11.0 & 12.4 & 2.5 & 6.0 & 4.1 & 2.1 & 13.2 & 7.6 & 6.3 & 11.2 \\
\cmidrule(lr){1-18}
CM-Align           & 74.5 & 50.2 & 62.5 & 19.3 & 14.9 & 14.6 & 18.4 & 39.7 & 18.3 & 4.4 & 16.7 & 24.5 & 19.0 & 44.4 & 24.6 & 12.5 & 17.4 \\
\midrule\midrule
\multicolumn{18}{l}{\textit{Qwen-2.5-7B}} \\
\midrule
Baseline           & 75.8 & 64.5 & 60.8 & 53.3 & 25.5 & 26.1 & 28.0 & 43.0 & 30.1 & 6.7 & 41.8 & 35.1 & 24.7 & 63.1 & 54.0 & 43.5 & 15.8 \\
Aligned            & 56.5 & 52.9 & 48.1 & 43.6 & 18.1 & 17.1 & 36.5 & 30.5 & 22.7 & 4.0 & 21.9 & 15.7 & 12.1 & 48.1 & 38.5 & 31.9 & 12.3 \\
\cmidrule(lr){1-18}
GRPO               & 78.4 & 70.2 & 70.2 & 64.9 & 31.4 & 34.3 & 50.7 & 47.2 & 30.9 & 14.6 & 43.9 & 41.0 & 48.1 & 64.8 & 55.2 & 49.8 & 44.4 \\
Aligned + GRPO     & 63.5 & 56.7 & 55.5 & 52.9 & 25.7 & 22.1 & 45.8 & 35.3 & 24.1 & 6.2 & 30.0 & 24.8 & 20.3 & 52.4 & 48.2 & 40.2 & 26.2 \\
\cmidrule(lr){1-18}
SFT                & 75.4 & 64.6 & 63.7 & 59.6 & 27.8 & 25.9 & 36.7 & 47.7 & 31.9 & 7.7 & 42.5 & 39.6 & 34.0 & 62.7 & 55.3 & 46.2 & 31.5 \\
Aligned + SFT      & 50.9 & 50.8 & 44.9 & 43.3 & 20.7 & 19.0 & 36.5 & 33.6 & 19.9 & 3.7 & 22.6 & 15.2 & 9.2 & 45.2 & 37.0 & 29.2 & 12.0 \\
\cmidrule(lr){1-18}
DCO                & 76.4 & 62.9 & 57.8 & 44.6 & 33.1 & 35.1 & 20.3 & 45.1 & 32.6 & 10.3 & 43.4 & 35.5 & 14.5 & 62.2 & 53.8 & 44.2 & 16.6 \\
Aligned + DCO      & 47.7 & 51.9 & 39.8 & 48.1 & 21.8 & 22.5 & 40.4 & 31.0 & 24.9 & 4.5 & 25.2 & 19.3 & 9.5 & 45.1 & 43.3 & 41.1 & 17.1 \\
\cmidrule(lr){1-18}
CM-Align           & 78.0 & 65.7 & 67.9 & 57.3 & 30.3 & 32.6 & 41.7 & 45.2 & 30.1 & 10.5 & 39.6 & 39.8 & 37.2 & 64.1 & 55.6 & 38.0 & 29.8 \\
\bottomrule
\end{tabular}
\caption{KLAR accuracy (\%) across languages, grouped into seen and held-out (unseen) languages.}
\label{tab:klar_appendix}
\end{table*}

\begin{table*}[t]
\centering
\scriptsize
\setlength{\tabcolsep}{3pt}
\begin{tabular}{lcccccccccccc}
\toprule
Model 
& en & de & es & fr & pt & id & zh & ar & ja & sw & bn 
& High \\
\midrule
\multicolumn{13}{l}{\textit{OLMo-2-1124-7B}} \\
\midrule
Baseline           & 66.25 & 49.75 & 49.75 & 47.00 & 48.75 & 43.25 & 45.00 & 38.00 & 39.00 & 30.25 & 32.00 & 47.94 \\
Aligned            & 66.25 & 47.50 & 46.50 & 45.75 & 46.50 & 42.75 & 43.75 & 33.25 & 40.00 & 30.50 & 27.50 & 46.19 \\
\cmidrule(lr){1-13}
GRPO               & 67.50 & 49.75 & 51.50 & 51.75 & 49.00 & 43.00 & 43.75 & 34.25 & 41.00 & 33.75 & 28.50 & 48.56 \\
Aligned + GRPO     & 63.50 & 49.50 & 47.75 & 49.50 & 48.25 & 44.25 & 43.25 & 33.25 & 41.00 & 33.00 & 29.25 & 47.00 \\
\cmidrule(lr){1-13}
SFT                & 65.00 & 50.25 & 49.00 & 53.00 & 48.75 & 45.75 & 45.25 & 33.50 & 39.25 & 30.75 & 25.75 & 48.00 \\
Aligned + SFT      & 65.50 & 50.00 & 49.75 & 53.00 & 48.25 & 44.00 & 44.50 & 29.50 & 37.75 & 30.50 & 25.50 & 47.28 \\
\cmidrule(lr){1-13}
DCO                & 67.25 & 49.25 & 52.00 & 48.25 & 49.00 & 44.00 & 45.75 & 39.00 & 39.00 & 30.00 & 29.75 & 48.69 \\
Aligned + DCO      & 66.00 & 47.50 & 50.75 & 51.25 & 49.75 & 44.00 & 42.50 & 36.00 & 40.75 & 34.75 & 30.25 & 48.06 \\
\cmidrule(lr){1-13}
CM-Align           & 66.75 & 51.25 & 51.00 & 51.00 & 47.75 & 45.00 & 39.00 & 33.50 & 39.75 & 28.50 & 27.50 & 47.50 \\
\midrule\midrule
\multicolumn{13}{l}{\textit{Qwen-2.5-7B}} \\
\midrule
Baseline           & 77.50 & 67.25 & 70.75 & 69.00 & 68.25 & 66.25 & 70.75 & 61.75 & 63.50 & 36.50 & 47.50 & 68.59 \\
Aligned            & 74.25 & 63.50 & 67.00 & 66.00 & 66.75 & 61.00 & 70.25 & 57.25 & 60.75 & 38.50 & 41.50 & 65.72 \\
\cmidrule(lr){1-13}
GRPO               & 76.00 & 66.75 & 70.50 & 68.00 & 68.00 & 65.50 & 69.50 & 60.25 & 64.75 & 36.50 & 42.00 & 67.97 \\
Aligned + GRPO     & 74.50 & 65.50 & 66.75 & 64.75 & 66.50 & 64.00 & 67.75 & 58.50 & 62.25 & 39.00 & 43.50 & 65.81 \\
\cmidrule(lr){1-13}
SFT                & 74.75 & 65.00 & 68.75 & 68.00 & 68.00 & 65.00 & 69.75 & 60.25 & 63.50 & 32.00 & 44.50 & 67.25 \\
Aligned + SFT      & 71.25 & 61.50 & 66.00 & 64.75 & 68.75 & 60.75 & 67.75 & 58.50 & 60.50 & 39.75 & 41.25 & 64.88 \\
\cmidrule(lr){1-13}
DCO                & 76.50 & 66.75 & 70.75 & 68.00 & 67.75 & 66.25 & 69.75 & 62.25 & 64.00 & 36.25 & 46.25 & 68.22 \\
Aligned + DCO      & 74.50 & 63.50 & 66.75 & 65.25 & 65.50 & 61.75 & 69.50 & 57.00 & 60.25 & 39.25 & 42.25 & 65.28 \\
\cmidrule(lr){1-13}
CM-Align           & 75.25 & 66.25 & 68.75 & 67.00 & 68.00 & 63.25 & 67.50 & 58.25 & 61.75 & 35.00 & 42.50 & 66.59 \\
\bottomrule
\end{tabular}
\caption{Global-MMLU accuracy (\%) across languages. High denotes the average over high-resource languages.}
\label{tab:global_mmlu_appendix}
\end{table*}

\end{document}